\documentclass[10pt,twocolumn,letterpaper]{article}

\usepackage[pagenumbers]{cvpr} %
\usepackage{calc}

\usepackage[dvipsnames]{xcolor}

\usepackage{epigraph}
\usepackage{algorithm}
\usepackage[noend]{algpseudocode}
\usepackage[misc]{ifsym}
\usepackage{graphicx} %

\renewcommand{\algorithmicrequire}{\textbf{Input:}}
\renewcommand{\algorithmicensure}{\textbf{Output:}}

\definecolor{cvprblue}{rgb}{0.21,0.49,0.74}
\usepackage[pagebackref,breaklinks,colorlinks,citecolor=cvprblue]{hyperref}

\def\paperID{*****} %
\def\confName{CVPR}
\def\confYear{2024}

\title{\textbf{\textit{\textcolor{black}{GoodDrag}}}: Towards Good Practices for Drag Editing with Diffusion Models}

\author{Zewei Zhang$^1$ \quad Huan Liu$^1$ \quad Jun Chen$^1$ \quad Xiangyu Xu$^2$\textsuperscript{\Letter}\\
$^1$McMaster University \quad $^2$Xi'an Jiaotong University\\
}

\begin{document}

\maketitle
\def\thefootnote{\Letter}\footnotetext{Research Lead, Corresponding Author.}\def\thefootnote{\arabic{footnote}}

\begin{abstract}
    In this paper, we introduce GoodDrag, a novel approach to improve the stability and image quality of drag editing. Unlike existing methods that struggle with accumulated perturbations and often result in distortions, GoodDrag introduces an AlDD framework that alternates between drag and denoising operations within the diffusion process, effectively improving the fidelity of the result. We also propose an information-preserving motion supervision operation that maintains the original features of the starting point for precise manipulation and artifact reduction. In addition, we contribute to the benchmarking of drag editing by introducing a new dataset, Drag100, and developing dedicated quality assessment metrics, Dragging Accuracy Index and Gemini Score, utilizing Large Multimodal Models. Extensive experiments demonstrate that the proposed GoodDrag compares favorably against the state-of-the-art approaches both qualitatively and quantitatively. 
    The project page is \url{https://gooddrag.github.io}.
\end{abstract}

\section{Introduction}
\label{sec:intro}

\begin{figure}[t]
\centering
\includegraphics[width=0.99\linewidth]{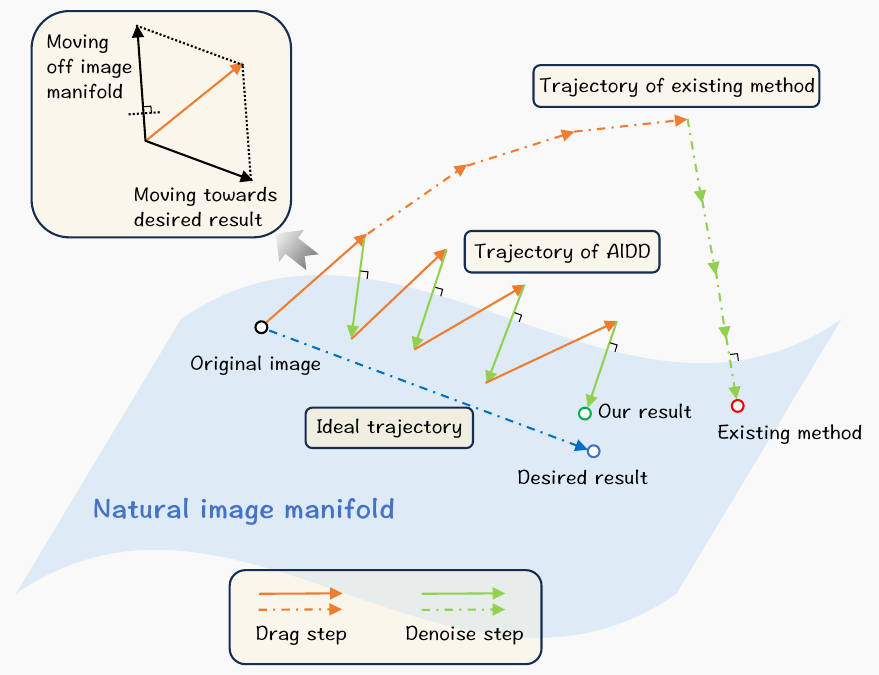}
\caption{ 
Existing diffusion-based drag editing methods (dotted trajectory), typically perform all drag operations at once, followed by denoising steps to correct the resulting perturbations. However, this approach often leads to accumulated perturbations that are too substantial for high-fidelity correction.
In contrast, the proposed AlDD framework (solid trajectory) alternates between drag and denoising operations within the diffusion process, effectively preventing the accumulation of large perturbations and ensuring more accurate editing results.
The drag operation modifies the image to achieve the desired dragging effect but introduces perturbations that deviate the intermediate result from the natural image manifold. 
The denoising operation, on the other hand, is trained to estimate the score function of the natural image distribution, guiding intermediate results back to the image manifold.
}
\vspace{-5mm}
\label{fig:teaser}
\end{figure}

\begin{figure*}
    \footnotesize\hspace{0em} Original \hspace{6.5em} User Edit \hspace{6em} GoodDrag \hspace{6em} Original \hspace{6em} User Edit \hspace{6em} GoodDrag
    \centering
    \includegraphics[width=1\linewidth]{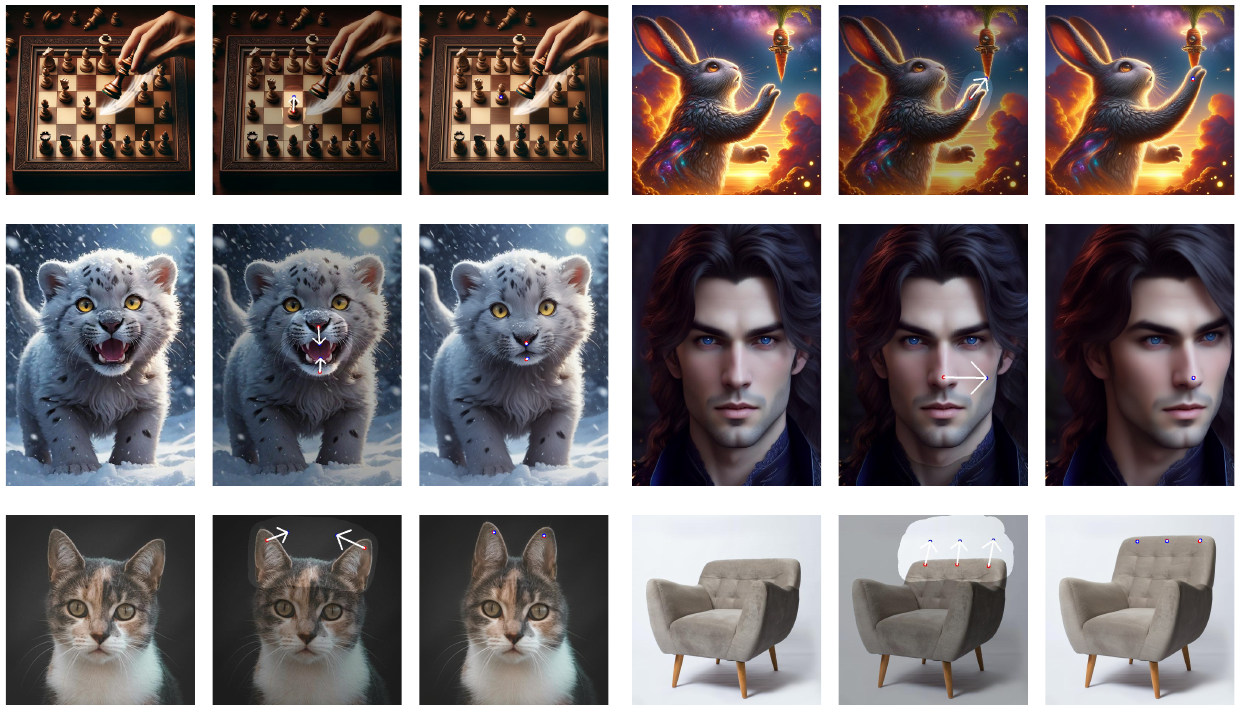}
    \caption{Given an input image (Original) and user-specified control points (User Edit), our proposed GoodDrag effectively ``drags'' the semantic contents from the initial handle point to the target point, as indicated by the white arrow. The blue point is the target point, fixed throughout the pipeline, while the red point represents the handle point moving closer to the target point during the optimization of GoodDrag. Optionally, users can select an indication mask to specify the editable region as shown in the User Edit column.
    }
\label{fig:show}
\end{figure*}

In this work, we present GoodDrag, a novel approach for drag editing with enhanced stability and image quality. 
Drag editing~\citep{pan2023_DragGAN} represents a new direction in generative image manipulation.
It allows users to intuitively edit images by specifying starting and target points, as if physically dragging an object or a part of an object from its initial location to the target location, with the edits blending harmoniously into the original image context as shown in Fig.~\ref{fig:show}.

Early methods~\citep{pan2023_DragGAN,ling2023freedrag} for drag editing employ Generative Adversarial Networks (GANs) \citep{goodfellow2014generative} that are often trained for class-specific images, and thereby struggle with generic, real-world images. 
Moreover, these methods rely heavily on GAN inversion techniques~\citep{roich2022pivotal,weihao2021gan,xu2023rigid}, which do not always work well for complex, in-the-wild scenarios.

To address these issues, recent advancements have shifted towards using diffusion models for drag editing~\citep{shi2023dragdiffusion,mou2023dragondiffusion,nie2023blessing}. Thanks to the remarkable capabilities of diffusion models in image generation, these methods have significantly improved the quality of drag editing for generic images. 
However, the current diffusion-based approaches often suffer from instability, which may result in outputs that have severe distortions or fail to adhere to designated control points.

This paper addresses these challenges by establishing two good practices for effective drag editing using diffusion models.
Our first contribution is Alternating Drag and Denoising (AlDD), a novel framework for diffusion-based drag editing. 
Existing methods typically conduct all drag operations at once and then attempt to correct the accumulated perturbations subsequently. However, this approach often leads to perturbations that are too substantial to be well-corrected. 
In contrast, the AlDD framework alternates between the drag and denoising operations within the diffusion process as shown in Fig.~\ref{fig:teaser}. This methodology effectively addresses the issue by preventing the accumulation of large distortions, ensuring a more refined and manageable editing process.

Our second contribution is the investigation into the artifacts in the edited results and the common failure of point control, where the starting point cannot be accurately dragged to the desired ending location. We identify the primary cause is that the dragged features in the existing algorithms could gradually deviate from the original features of the starting point. 
To tackle this issue, we propose an information-preserving motion supervision operation that maintains the original features of the starting point, ensuring realistic and precise point manipulation.

Furthermore, we make early efforts to benchmark drag editing by introducing a new dataset along with dedicated evaluation metrics. Notably, we develop Gemini Score, a novel quality assessment metric utilizing Large Multimodal Models~\citep{team2023gemini}, which is more reliable and effective than existing No-Reference Image Quality Assessment metrics.

Combining these good practices, our final algorithm, named GoodDrag, consistently achieves high-quality results for drag editing as shown in Fig.~\ref{fig:show}. 
Extensive experiments demonstrate the effectiveness of GoodDrag, outperforming state-of-the-art approaches both quantitatively and qualitatively.
\section{Related Work} 
\label{sec:related}

\subsection{Diffusion-Based Image Manipulation}
In image editing tasks such as inpainting, colorization, and text-driven editing, GANs have been extensively utilized~\citep{xu2017learning,yu2018generative,isola2017image,park2019semantic,su2023styleretoucher,liu2023gan, lezama2022discrete,chen2020deepfacedrawing,chen2021deepfaceediting,du2023one}.
While these methods have shown the ability to edit both generated and real images~\citep{roich2022pivotal}, they are often constrained by the limitations of GANs, such as restricted content range in edited images and suboptimal image quality. In contrast, the diffusion models ~\citep{sohl2015deep,ho2020denoising, song2020denoising, song2020score, rombach2022high, su2022drawinginstyles, yan2024training} offer more flexibility in control conditions for image generation and editing. They produce higher quality results across a broader range of images compared to GANs~\citep{dhariwal2021diffusion}. This advancement allows for more nuanced and detailed manipulations, significantly enhancing the scope and fidelity of image editing.

Recently, diffusion models have been extensively used in image manipulation and generation~\citep{lin2023text, gupta2023photorealistic}. In inpainting task, diffusion models can generate high-quality content~\citep{saharia2022palette, nichol2021glide} and can also incorporate additional conditions.
Diffusion models are applied not only in general image restoration~\citep{kawar2022denoising} but also in specific scenarios like restoring images affected by weather conditions such as rain and snow~\citep{ozdenizci2023restoring}.
Diffusion models are not only suited for various image editing tasks but also accommodate flexible control inputs.
For instance, the Dreambooth series~\citep{ruiz2023dreambooth, raj2023dreambooth3d, ruiz2023hyperdreambooth} uses a set of images with the same theme to edit and create new content within that theme. 
CustomSketching~\citep{xiao2024customsketching} leverages sketches and text to guide the generation of images.
Meanwhile, ControlNet~\citep{zhang2023adding} offers more flexible control methods, such as those based on the canny edge, user scribbles, and more.
As mentioned above, diffusion models have proven their practicality in a wide range of image editing tasks, consistently producing high-quality results.

\subsection{Drag Editing}
Drag editing, first introduced in DragGAN~\citep{pan2023_DragGAN}, represents an innovative technique in the field of image editing. This approach allows users to interactively, intuitively, and dynamically alter the content of an image. By simply specifying a starting and an ending point within the image, drag editing enables users to achieve complex modifications with relative ease. However, subsequent updates, as noted in ~\citep{ling2023freedrag}, have pointed out some instabilities in DragGAN, deviating from the intended drag tasks, and proposed a more stable method.
Nevertheless, these methods are inherently reliant on GANs models. This dependence means that they cannot be directly applied to user-input images but are limited to images generated by GANs.
Employing ~\citep{roich2022pivotal} enables the specification of particular GANs models for drag editing on the output images. 
However, this approach, dependent on pre-trained GANs models, has its limitations. It may not be feasible for certain types of images, such as those featuring rare or less common subjects like specific animal species.
Moreover, images containing a mix of different object types may not be suitable for GANs models. Consequently, these GANs-based drag editing methods ~\citep{pan2023_DragGAN, ling2023freedrag} face practical limitations when applied to general user-input images, hindering their ability to perform drag editing tasks across a broad spectrum of scenarios.

To overcome the limitations of GAN-based drag editing, \citep{shi2023dragdiffusion, nie2023blessing} have successfully integrated this technique with diffusion models. Thanks to the capabilities of diffusion models \citep{sohl2015deep, ho2020denoising, song2020denoising, song2020score, rombach2022high}, coupled with the rapid training facilitated by LoRA \citep{hu2021lora}, it is now feasible to perform drag editing on any image while substantially preserving the details of the original image. 
However, these diffusion-based methods exhibit instability, occasionally resulting in outputs of lower image quality. This instability is partly due to the broader range of image sources, presenting greater challenges in drag editing. Additionally, diffusion models typically edit within the generative process of the same image, unlike GAN-based methods that generate a new image at each drag edit step.
This accumulated editing can lead to artifacts, compromising the stability of the final image.

In response to these issues, we propose the Alternating-Drag-and-Denoising (AlDD) framework. AlDD disperses the impact of drag editing throughout the image generation process, enabling changes to evolve progressively rather than accumulating at a specific generative stage. We also introduce an information-preserving method of drag editing, which mitigates the feature drifting and stabilizes the overall diffusion process for image generation. This approach ensures the production of high-quality images in drag editing, effectively addressing the challenges posed by previous methods.

\section{Method}
\label{sec:method}
In this work, we propose GoodDrag, a new framework, for high-quality drag editing with diffusion models~\citep{song2020denoising, song2020score, rombach2022high}.
We develop and integrate two effective practices within this framework: Alternate Drag and Denoising (Section~\ref{sec:AlDD}) and Information-Preserving Motion Supervision (Section~\ref{sec:info_preserve}), which are instrumental in reducing visual artifacts and enhancing precision in drag editing. 

\begin{figure*}[htbp]
    \centering
    \includegraphics[width=\textwidth]{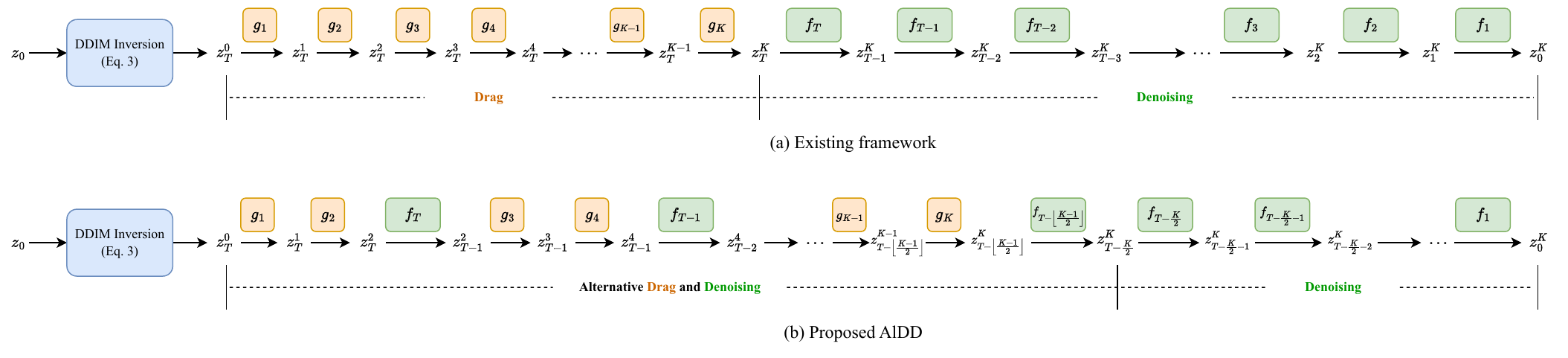}
    \caption{Overview of the proposed AlDD framework. 
    (a) Existing methods first perform all drag editing operations $\{g_k\}_{k=1}^K$ at a single time step $T$ and subsequently apply all denoising operations $\{f_t\}_{t=T}^1$ to transform the edited image $z_T^K$ into the VAE image space. 
    (b) To mitigate the accumulated perturbations in (a), AlDD alternates between the drag operation $g$ and the diffusion denoising operation $f$, which leads to higher quality results. Specifically, we apply one denoising operation after every $B$ drag steps and ensure the total number of drag steps $K$ is divisible by $B$. 
    We set $B=2$ in this figure for clarity.
    }
    \label{fig:AlDD}
\end{figure*}

\subsection{Preliminary on Diffusion Models}
Diffusion models represent a compelling subclass of generative models, having demonstrated remarkable performance in synthesizing high-quality images, as evidenced by advanced applications like DALLE2~\citep{ramesh2022hierarchical} and Stable Diffusion~\citep{rombach2022high}. 
These models consist of two distinct phases: the forward process and the reverse process.

In the forward process, a given data sample $z_0$ is combined with increasing levels of Gaussian noise over a series of $T_{\text{max}}$ steps. 
This process results in the generation of a series of progressively noised samples $\{z_t\}_{t=1}^{T_{\text{max}}}$, with each $z_t$ representing the noised image at time step $t$. 
Mathematically, the forward process can be formulated as:
\begin{align}\label{eq:forward}
z_t = \sqrt{\alpha_t}z_{0} + \sqrt{1-\alpha_t} \varepsilon,
\end{align}
where $\varepsilon \sim \mathcal{N}(0, \mathbf{I})$ is a random Gaussian noise. 
$\alpha_t \in (0,1)$ acts as a diminishing factor of $z_0$, and the sequence $\{\alpha_t\}_{t=1}^{T_{\text{max}}}$ is designed to be monotonically decreasing for a stronger diminishing effect and a stronger noise as $t$ increases.
$\alpha_{T_{\text{max}}}$ is close to 0, and $z_{T_{\text{max}}}$ approximates an isotropic Gaussian distribution.

During the reverse process, we first sample $z_{T_{\text{max}}}$ from the standard Gaussian distribution $\mathcal{N}(0,\mathbf{I})$ and then generate samples resembling the original data distribution of $z_0$ by gradually reducing the noise levels.
The Denoising Diffusion Implicit Models (DDIM)~\citep{song2020denoising} stand out in this phase, achieving decent efficiency and consistency in generating high-quality images. 
The reverse process from $z_t$ to $z_{t-1}$ under the deterministic DDIM framework can be written as:
\begin{equation}\label{eq:reverse}
    z_{t-1}=\sqrt{\alpha_{t-1}}\frac{z_t-\sqrt{1-\alpha_t}\varepsilon_{\theta}(z_t, t)}{\sqrt{\alpha_t}}+\sqrt{1-\alpha_{t-1}}\varepsilon_{\theta}(z_t, t),
\end{equation}
where $\varepsilon_{\theta}$ represents a neural network with parameters $\theta$, which is trained to predict the noise $\varepsilon$ in Eq.~\ref{eq:forward}.
For clarity, we denote Eq.~\ref{eq:reverse} as $z_{t-1}=f_t(z_t)$.

\vspace{1mm}
\noindent\textbf{DDIM Inversion.} 
The deterministic nature of DDIM allows the transformation of a natural image $z_0$ to its latent variable ${z}_t$ (the inverse operation of Eq.~\ref{eq:reverse}). 
As suggested in~\citep{song2020denoising}, the inversion from ${z}_{t-1}$ to ${z}_{t}$ is formulated as:
\begin{equation}\label{eq:inversion}
\begin{split}
    {z}_{t} = & \sqrt{\alpha_{t}}\left( \sqrt{\dfrac{1}{\alpha_{t}}-1}-\sqrt{\dfrac{1}{\alpha_{t-1}}-1}\right)\cdot \varepsilon_{\theta}({z}_{t-1}, t-1)
    \\&+ \sqrt{\dfrac{\alpha_{t}}{\alpha_{t-1}}}{z}_{t-1},
\end{split}
\end{equation}
which can be directly derived from Eq.~\ref{eq:reverse}, where $\varepsilon_{\theta}(z_{t-1}, t-1)$ is used to approximate $\varepsilon_{\theta}(z_{t}, t)$.
The DDIM inversion is invaluable for image editing applications, where one can apply targeted modifications to the latent variable $z_t$ and then transform the edited latent variable back to the image space by denoising with Eq.~\ref{eq:reverse}.
This circumvents the difficulties of directly modifying $z_0$, enabling more flexible and practical image editing applications.

Following Stable Diffusion~\citep{rombach2022high}, we use the Variational Autoencoder (VAE)~\citep{esser2021taming} to encode original images into lower-resolution images in feature space to reduce computation and memory costs. 
Throughout the paper, the variables denoted by $z$ refer to images in this VAE space instead of the pixel space.

\subsection{Drag Editing} \label{sec:dragediting}

The input of drag editing is a source image $z_0$, a set of $l$ starting points $\{\boldsymbol{p}_i \}$, and their corresponding target points $\{\boldsymbol{q}_i\}$, where $i=1,2,\cdots,l$.
Here, $\boldsymbol{p}_i, \boldsymbol{q}_i \in \mathbb{R}^2$ represent 2D pixel coordinates within the image plane.
An optional binary mask $\rm{M}$ can also be provided to specify the image region that is allowed for edits.
The objective of drag editing is to seamlessly transfer content from each starting point $\boldsymbol{p}_i$ to the designated target point $\boldsymbol{q}_i$, while ensuring that the resulting image remains natural and cohesive, with the edits blending harmoniously into the original image context.

The drag editing starts by transforming the source image $z_0$ into a latent representation $z_T$ through the DDIM inversion (Eq.~\ref{eq:inversion}), where the timestep $T$ is empirically chosen, typically close to $T_\text{max}$.
With the transformed $z_T$, the input image can be edited through a $K$-step iterative process as shown in Fig.~\ref{fig:AlDD}(a).
Each iteration, denoted by $g_k$, $k=1, \cdots, K$, comprises two main phases: motion supervision and point tracking~\citep{pan2023_DragGAN,shi2023dragdiffusion}.

\vspace{1mm}
\noindent\textbf{Motion supervision.} 
We denote the output of the $k$-th iteration, which serves as the input for the $(k+1)$-th iteration, as $z_T^k$ and the corresponding handle points as $\boldsymbol{p}_i^k$, with the initial image $z_T^0=z_T$ and the initial handle point $\boldsymbol{p}_i^0 = \boldsymbol{p}_i$.
The aim of motion supervision is to progressively edit the current image $z_T^k$ to move the handle points $\boldsymbol{p}_i^k$ towards their targets $\boldsymbol{q}_i$.

Specifically, denoting the movement direction for the $i$-th point as $\boldsymbol{d}_i^k = \frac{\boldsymbol{q}_i-\boldsymbol{p}_i^k}{\|{\boldsymbol{q}_i-\boldsymbol{p}_i^k}\|_2}$,
the motion supervision is realized by aligning the feature of $z_T^k$ around point $\boldsymbol{p}_i^k+ \beta \boldsymbol{d}_i^k$ to the feature around $\boldsymbol{p}_i^k$, where $\beta$ is the step size of the movement.
The feature of $z_T^k$ can be written as $\mathrm{F}(z_T^k) = \mathcal{I}\left( \mathrm{U}_\theta(z_T^k;T) \right)$, where the feature extractor $\mathrm{U}_\theta$ is the U-Net of Stable Diffusion parameterized by $\theta$, and $\mathcal{I}$ represents the interpolation function to adjust the feature map to the size of the input image. 
The feature alignment is captured by the following loss function:
\begin{equation}
\label{eq:base_motion_loss}
\footnotesize
\begin{aligned}
\mathcal{L}(z_T^k; \{\boldsymbol{p}_i^k\}) = &\sum_{i=1}^l \left\Vert \mathrm{F}_{\mathrm{\Omega}(\boldsymbol{p}_i^k+\beta \boldsymbol{d}_i^k, r_1)}(z_T^k) - \text{sg}\left( \mathrm{F}_{\mathrm{\Omega}(\boldsymbol{p}_i^k, r_1)}(z_T^k) \right) \right\Vert_1 \\
&+ \lambda\left\Vert \left(z_{T-1}^k-{\text{sg}}\left(z_{T-1}^0\right)\right)\odot (1-\rm{M})\right\Vert_1,
\end{aligned}
\end{equation}
where ${\mathrm{\Omega}}(\boldsymbol{p}_i^k, r_1) = \{\boldsymbol{p} \in \mathbb{Z}^2 : \|\boldsymbol{p} - \boldsymbol{p}_i^k \|_\infty \leqslant r_1\}$ describes a square region centered at $\boldsymbol{p}_i^k$ with a radius $r_1$.
$\text{sg}(\cdot)$ denotes the stop-gradient operation.
The first term of Eq.~\ref{eq:base_motion_loss} essentially drives the appearance of the image around $\boldsymbol{p}_i^k+ \beta \boldsymbol{d}_i^k$ to get closer to the appearance around $\boldsymbol{p}_i^k$.
The second term ensures the non-editable region, as indicated by $1-\rm{M}$, remains unchanged throughout the editing process.

Finally, the motion supervision for the $(k+1)$-th iteration takes one gradient descent step according to the feature alignment loss $\mathcal{L}(z_T^k; \{\boldsymbol{p}_i^k\})$:
\begin{equation}
\label{eq:base_grad_descent}
    z_T^{k+1}  = z_T^k - \eta \cdot \frac{\partial \mathcal{L}(z_T^k; \{\boldsymbol{p}_i^k\})}{\partial z_T^k},
\end{equation}
where $\eta$ is the step size.

\vspace{1mm}
\noindent\textbf{Point tracking.}
While the motion supervision effectively guides the movement of the handle point towards $\boldsymbol{p}_i^k+\beta \boldsymbol{d}_i^k$, its final position at this exact spot is not guaranteed.
This necessitates the point tracking to locate the new location of the handle point $\boldsymbol{p}_i^{k+1}$, which is formulated as:
\begin{equation}
\label{eqn:point_tracking}
\boldsymbol{p}_i^{k+1} = \operatorname*{argmin}_{\boldsymbol{p} \in {\mathrm{\Omega}}(\boldsymbol{p}_i^k,r_2)} \left\Vert \mathrm{F}_{\boldsymbol{p}}(z_T^{k+1}) - \mathrm{F}_{\boldsymbol{p}_i^0}(z_T^0)\right\Vert_1.
\end{equation}

Eq.~\ref{eqn:point_tracking} identifies the updated handle point by searching the location in $z_T^{k+1}$ that most closely resembles the original starting point $\boldsymbol{p}_i^0$ in the original image $z_T^0$ based on feature similarity.
$r_2$ denotes the radius of the search area ${\mathrm{\Omega}}(\boldsymbol{p}_i^k, r_2)$.

\vspace{1mm}
\noindent\textbf{Iterative editing.}
We represent Eq.~\ref{eq:base_grad_descent} as $z_T^{k+1}  = g_{k+1}(z_T^k)$.
It is worth noting that Eq.~\ref{eqn:point_tracking} is also involved in Eq.~\ref{eq:base_grad_descent} which is dependent on the tracking of the handle point $\boldsymbol{p}_i^{k}$ (the dependence is omitted in $f$ for simplicity). 

As shown in Fig.~\ref{fig:AlDD}(a), the editing process begins by sequentially performing the drag operations $\{g_k\}_{k=1}^K$ in the latent space $z_T$.
The resulting image $z_T^K$ is transitioned back to the VAE image space by applying the denoising operations $\{f_t\}_{t=T}^1$ as described by Eq.~\ref{eq:reverse}. 
The final output is $\hat{z}_0 = z_0^K$.

\subsection{Alternating Drag and Denoising}
\label{sec:AlDD}

\begin{figure}[t]
	 \centering
        \includegraphics[width=1\linewidth]{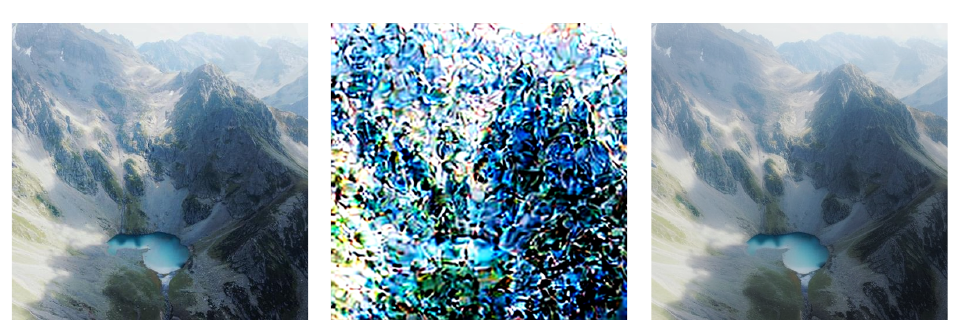}
        \vspace{-1mm}
        \hspace{1em} \scriptsize (a) Original \hspace{4em} (b) Single time step\hspace{3em} (c) Multiple time steps
	\caption{We generate 10 random noise samples from the distribution $\mathcal{N}(0,0.1^2\mathbf{I})$ and compare two scenarios: (b) adding all samples simultaneously to $z_T$ and (c) adding each sample individually across 10 different time steps. In the former case, where all noise samples are added to $z_T$ at once, the resulting image exhibits significant degradation. In contrast, when we distribute the noise samples across multiple time steps, the resulting image well preserves the original content with high fidelity.}
	\label{fig:AlDD toy example}
\end{figure}

\epigraph{\textit{"A stitch in time saves nine."}}{--- Proverb}

While existing drag editing methods~\citep{shi2023dragdiffusion,mou2023dragondiffusion} have achieved promising results, they inherently suffer from low fidelity. 
This issue mainly stems from the heuristic nature of the drag operation, which introduces undesirable perturbation to $z_T$ during the feature alignment in Eq.~\ref{eq:base_motion_loss}.
While subsequent denoising operations aim to rectify these perturbations, performing all the drag operations within a single diffusion time step leads to accumulated perturbations and distorations that are too substantial for accurate correction.

To address this challenge, we propose a novel framework for drag editing with diffusion models, termed Alternating Drag and Denoising (AlDD).
The core of AlDD lies in distributing editing operations across multiple time steps within the diffusion process.
It involves alternating between drag and denoising steps, allowing for more manageable and incremental changes. 
As illustrated in Fig.~\ref{fig:AlDD}(b), after applying $B$ drag operations $g$ at time step $t$, a denoising step $f$ follows, which alleviates the undesirable artifacts introduced by feature alignment by converting the latent representation from $t$ to $t-1$. 
We then perform the subsequent $B$ drag operations on time step ${t-1}$, and this pattern continues until all intended drag edits are completed. 
The feature alignment loss for motion supervision in AlDD is defined as:
\begin{equation}
\label{eq:aldd_motion_loss}
\footnotesize
\begin{aligned}
\mathcal{L}(z_t^k; \{\boldsymbol{p}_i^k\}) = &\sum_{i=1}^l \left\Vert \mathrm{F}_{\mathrm{\Omega}(\boldsymbol{p}_i^k+\beta \boldsymbol{d}_i^k, r_1)}(z_t^k) - \text{sg}\left( \mathrm{F}_{\mathrm{\Omega}(\boldsymbol{p}_i^k, r_1)}(z_t^k) \right) \right\Vert_1 \\
&+ \lambda\left\Vert \left(z_{t-1}^k-{\text{sg}}\left(z_{t-1}^0\right)\right)\odot (1-\rm{M})\right\Vert_1.
\end{aligned}
\end{equation}
In this equation, since the image $z_t^k$ has undergone $\left\lfloor \frac{k}{B} \right\rfloor$ denoising operations, we apply the drag operation at the diffusion time step $t=T- \left\lfloor \frac{k}{B} \right\rfloor$. 
This is in sharp contrast to Eq.~\ref{eq:base_motion_loss}, which applies all drag operations at a single time step $T$.

Finally, we conduct the remaining denoising steps to convert the latent representation to the desired VAE image space $z_0$. 
Notably, the AlDD only changes the order of the computations, which improves editing quality without introducing additional computational overhead.

The key insight behind this framework is that addressing perturbations incrementally as they arise, rather than allowing them to accumulate, facilitates more effective and manageable image editing. 
In other words, it is better to fix the problem when it is small than to wait until it becomes more significant.

To validate this concept, we conduct a toy experiment as shown in Fig.~\ref{fig:AlDD toy example}. 
We simulate the perturbations introduced during image editing with random Gaussian noise, and compare the results of adding multiple noise samples within the same diffusion time step versus across different time steps. 
When noise is added all at once to $z_T$, the resulting image suffers from low fidelity as shown in Fig.~\ref{fig:AlDD toy example}(b). 
This is due to the accumulation of noise within a single time step, leading to a substantial deviation from the image manifold (Fig.~\ref{fig:teaser}). 
In contrast, distributing the noise across multiple diffusion steps results in well-corrected perturbations and better preservation of original content, as shown in Fig.~\ref{fig:AlDD toy example}(c). 
This validates our hypothesis that progressive adjustments lead to more effective image editing.
Further analysis and results of AlDD are presented in Section~\ref{sec:analysis}.

\subsection{Information-Preserving Motion Supervision}\label{sec:info_preserve}

\begin{figure}[t]
\centering
      \includegraphics[width=0.5\textwidth]{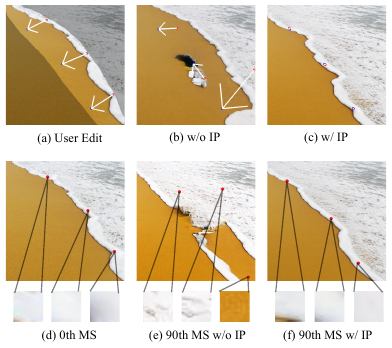}

    \caption{Illustration of the feature drifting issue. In (d), the initial handle points are located near the boundary of the beach wave. As drag editing progresses, the features of the handle points deviate from their original appearance. We show the intermediate result at the 90th motion supervision (MS) step in (e), where the handle points have drifted away from the wave boundary, leading to artifacts and inaccurate point movement in (b). 
    To alleviate this issue, we propose information-preserving motion supervision (IP) to preserve the fidelity of the handle points to the original points as shown in (f), which effectively facilitates higher-quality results in (c).
    }
    \label{fig:ip-insight}
\end{figure}

Another challenge in existing drag editing methods is the feature drifting of handle points, which can lead to artifacts in the edited results and failures in accurately moving handle points as shown in Fig.~\ref{fig:ip-insight}(b). 
The feature drifting issue is illustrated in the second row of Fig.~\ref{fig:ip-insight}, where the initial handle points (red points) in Fig.~\ref{fig:ip-insight}(d) are near the boundary of the beach wave. 
As the number of drag steps increases, the handle points become less similar to their original appearance, drifting away from the wave boundary towards the sea foam or the sand, as shown in Fig.~\ref{fig:ip-insight}(e).

We identify that the root cause of handle point drifting lies in the design of the motion supervision loss, as defined in Eq.~\ref{eq:base_motion_loss}. 
This loss function encourages the next handle point, $\boldsymbol{p}_i^k+\beta \boldsymbol{d}_i^k$, to be similar to the current handle point, $\boldsymbol{p}_i^k$. Consequently, even minor drifts in one iteration can accumulate over time during motion supervision, leading to significant deviations and distorted outcomes.

To address this problem, we propose an information-preserving motion supervision approach, which maintains the consistency of the handle point with the original point throughout the editing process. 
The updated feature alignment loss for motion supervision is formulated as:
\begin{equation}
\label{eq:new_motion_loss}
\footnotesize
\begin{aligned}
\mathcal{L}(z_t^k; \{\boldsymbol{p}_i^k\}) = &\sum_{i=1}^l \left\Vert \mathrm{F}_{\mathrm{\Omega}(\boldsymbol{p}_i^k+\beta \boldsymbol{d}_i^k, r_1)}(z_t^k) - \text{sg}\left( \mathrm{F}_{\mathrm{\Omega}(\boldsymbol{p}_i^0, r_1)}(z_t^0) \right) \right\Vert_1 \\
&+ \lambda\left\Vert \left(z_{t-1}^k-{\text{sg}}\left(z_{t-1}^0\right)\right)\odot (1-\rm{M})\right\Vert_1,
\end{aligned}
\end{equation}
where $\boldsymbol{p}_i^0$ is the original handle point in the unedited image $z_t^0$. This formulation ensures that the intended handle point $\boldsymbol{p}_i^k+\beta \boldsymbol{d}_i^k$ in the edited image $z_t^k$ remains faithful to the original handle point, thereby preserving the integrity of the editing process.

While the information-preserving motion supervision effectively addresses the handle point drifting issue, it introduces new challenges.
Specifically, Eq.~\ref{eq:new_motion_loss} is more difficult to optimize due to its typically larger feature distance than the original motion supervision loss Eq.~\ref{eq:base_motion_loss}.
Therefore, a straightforward application of Eq.~\ref{eq:new_motion_loss} often results in unsuccessful dragging effects of the handle point. 
Initially, we attempted to overcome this by increasing the step size $\eta$ in the motion supervision process (Eq.~\ref{eq:base_grad_descent}), which turned out to be less effective.
Instead, we find that maintaining a small step size and increasing the number of motion supervision steps before each point tracking offers a better solution:
\begin{equation}
\label{eq:new_grad_descent}
z_{t,j+1}^{k} = z_{t,j}^{k} - \eta \cdot \frac{\partial \mathcal{L}(z_{t,j}^{k}; \{\boldsymbol{p}_i^k\})}{\partial z_{t,j}^k}, ~~ j=0,\cdots,J-1,
\end{equation}
where $z_{t,0}^{k} = z_{t}^{k}$ is the initial image, and $z_{t}^{k+1} = z_{t,J}^{k}$ is the output after $J$ gradient steps.

The proposed information-preserving motion supervision marks an effective practice for drag editing, which ensures that the handle point remains close to its original appearance without introducing excessive artifacts as shown in Fig.~\ref{fig:ip-insight}(f). Consequently, this leads to higher-quality results, as evidenced in Fig.~\ref{fig:ip-insight}(c). 
It is worth noting that although the proposed solution appears simple, its development demands a deep understanding of the underlying problem and meticulous engineering efforts.

Finally, the whole pipeline of GoodDrag is summarized in Algorithm~\ref{alg:gooddrag}.
Similar to DragDiffusion~\citep{shi2023dragdiffusion}, we also use LoRA~\citep{hu2021lora} to finetune the diffusion U-Net for better denoising performance with Stable Diffusion~\citep{rombach2022high}.

\begin{algorithm}[!t]
\caption{Pipeline of GoodDrag}
    \algorithmicrequire{
        Input image $z_0$, binary mask for editable region ${\rm M}$, handle points $\{\boldsymbol{p}_{i}\}_{i=1}^l$, target points $\{\boldsymbol{q}_{i}\}_{i=1}^l$, U-Net $\mathrm{U}_{\theta}$, latent time step $T$, number of drag iterations $K$, number of motion supervision steps per point tracking $J$ \\
        }
        \algorithmicensure{
        Output image $\hat{z}_0$}
         
         \begin{algorithmic}[1]
         \State Finetune $\mathrm{U}_{\theta}$ on $z_0$ with LoRA
         \State $z_T \gets$ apply DDIM inversion to $z_0$ (Eq.~\ref{eq:inversion}) 
         \State $z_T^0 \gets z_T$, $\boldsymbol{p}_{i}^0 \gets \boldsymbol{p}_{i}$
         \For{$k$ in $0:K-1$}
            \State $t = T-\left\lfloor \frac{k}{B} \right\rfloor$
            \State $z_{t,0}^k \gets z_t^k$
            \For{$j$ in $0:J-1$}
                \State $\mathrm{F}(z_{t,j}^k) \gets \mathcal{I}\left(\mathrm{U}_{\theta}(z_{t,j}^k; t)\right)$
                \State Update $z_{t,j+1}^k$ using motion supervision as Eq.~\ref{eq:new_grad_descent}
            \EndFor
            \State $z_t^{k+1} \gets z_{t,J}^k$
            \State Update $\{\boldsymbol{p}_{i}^{k+1}\}_{i=1}^l$ using points tracking as Eq.~\ref{eqn:point_tracking}
            \If{$(k+1) \bmod B=0$}
                \State $z_{t-1}^{k+1} \gets$ one step denoising from $z_{t}^{k+1}$ with Eq.~\ref{eq:reverse}
            \EndIf
         \EndFor
         \For{$t$ in $T- \frac{K}{B}:1$}
            \State $z_{t-1}^{K} \gets$ one step denoising from $z_{t}^{K}$ with Eq.~\ref{eq:reverse}
         \EndFor
         \State $\hat{z}_0 \gets z_0^K$
         \end{algorithmic}
         \label{alg:gooddrag}
\end{algorithm}

\section{Benchmark}
\label{sec:benchmark}

To benchmark the progress in drag-based image editing, we introduce a new evaluation dataset named Drag100, and two dedicated quality assessment metrics, DAI and GScore.

\subsection{Drag100 Dataset}
\begin{figure*}[!htbp]
    \centering
    \includegraphics[width=1\linewidth]{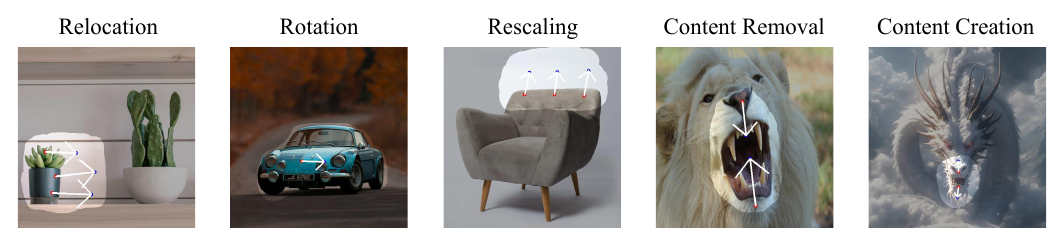}
    \caption{Example images and user edits from the Drag100 benchmark.}
    \label{fig:Dataset tasks examples}
\end{figure*}

\begin{figure}[t]
    \centering
    \includegraphics[width=1\linewidth]{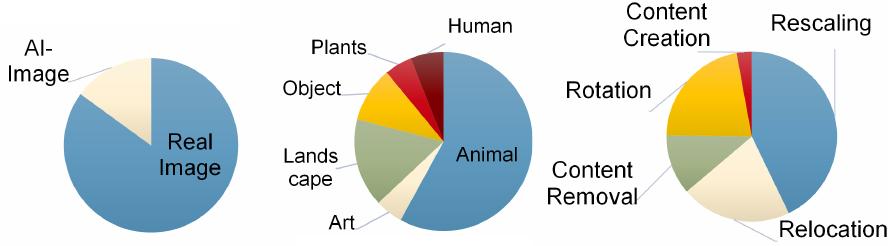}
    \caption{Distribution of various categories and tasks in the Drag100 dataset.}
    \label{fig:Dataset tasks}
\end{figure}

Since drag-based image editing is still a nascent research area, there is a lack of evaluation datasets. 
While recent works have introduced two datasets~\citep{shi2023dragdiffusion,nie2023blessing}, they have certain limitations. 
First, they do not provide indication masks $\text{M}$ for drag editing, and thus each algorithm can freely choose its own masks. 
Since different masks may give inconsistent results, this limitation can lead to uncontrolled experiments and difficulties in benchmarking and fair comparison of different methods. 
Second, these datasets were not constructed with explicit consideration for diversity, making evaluations less comprehensive.

To overcome these challenges, we introduce a new dataset called Drag100. This dataset consists of 100 images, each with carefully labeled masks and control points, ensuring that different methods can be evaluated in a controlled manner. Fig.~\ref{fig:Dataset tasks examples} showcases some examples from Drag100.

Drag100 is particularly designed to encompass a diverse range of content, as shown in Fig.~\ref{fig:Dataset tasks}. 
It comprises 85 real images and 15 AI-generated images using Stable Diffusion. 
The dataset spans various categories, including 58 animal images, 5 artistic paintings, 16 landscapes, 5 plant images, 6 human portraits, and 10 images of common objects such as cars and furniture.

We have also considered the diversity of drag tasks, including relocation, rotation, rescaling, content removal, and content creation, as illustrated in Fig.~\ref{fig:Dataset tasks examples}. 
These tasks have distinct characteristics. 
Relocation involves moving an object or a part of an object, while rotation adjusts the orientation of objects; both tasks primarily focus on the ability to mimic rigid motion in the physical world without changing the object area or creating new contents.
Rescaling corresponds to enlarging or shrinking an object, typically affecting its size. 
Content removal involves deletion of specific image components, \textit{e.g.}, closing mouth, whereas content creation involves generating new content not present in the original image, \textit{e.g.}, opening mouth. 
These tasks often have a higher requirement for hallucination capabilities, similar to occlusion removal~\citep{liu2020learning} and image inpainting~\citep{yu2018generative}.
By including these diverse settings, the Drag100 dataset facilitates a comprehensive evaluation of various aspects of drag editing algorithms.

\subsection{Evaluation Metrics for Drag Editing}\label{sec:metrics}
In this work, we introduce the following two quality assessment metrics, Dragging Accuracy Index (DAI) and Gemini Score (GScore), for quantitative evaluation.

\vspace{1mm}
\noindent\textbf{DAI.}
We introduce DAI to quantify the effectiveness of an approach in transferring the semantic contents to the target point.
In other words, the objective of DAI is to assess whether the source content at $\boldsymbol{p}_i$ of the original image has been successfully dragged to the target location $\boldsymbol{q}_i$ in the edited image.
Mathematically, the DAI is defined as:
\begin{equation} \label{eq:DAI}
    {\rm DAI} =  \dfrac{1}{l} \sum_{i=1}^{l} \dfrac{\left\Vert {\phi(z_0)_{\mathrm{\Omega}(\boldsymbol{p}_i,\gamma)}} - \phi(\hat{z}_0)_{{\rm\Omega}(\boldsymbol{q}_i,\gamma)}\right\Vert_2^2}{(1+2\gamma)^2},
\end{equation}
where $\phi$ is the VAE decoder converting $z_0$ to the RGB image space, and $\mathrm{\Omega}(\boldsymbol{p}_i,\gamma)$ denotes a patch centered at $\boldsymbol{p}_i$ with radius $\gamma$.
Eq.~\ref{eq:DAI} calculates the mean squared error between the patch at $\boldsymbol{p}_i$ of $\phi({z}_0)$ and the patch at $\boldsymbol{q}_i$ of $\phi(\hat{z}_0)$. 
By varying the radius $\gamma$, we can flexibly control the extent of context incorporated in the assessment: a small $\gamma$ ensures precise measurement of the difference at the control points, while a large $\gamma$ encompasses a broader context; this serves as a lens to examine different aspects of the editing quality. 

\vspace{1mm}
\noindent\textbf{GScore.}
While the proposed DAI is effective in measuring drag accuracy, it alone is not sufficient as the editing process could introduce distortions or artifacts, resulting in unrealistic outcomes. 
Therefore, evaluating the naturalness and fidelity of the edited images is important to ensure a comprehensive quality assessment.

This evaluation is particularly challenging as there is no ground-truth image available for reference.
Existing No-Reference Image Quality Assessment (NR-IQA) methods, such as~\citep{ke2021musiq,golestaneh2022no,chen2023topiq}, offer a way to assess image quality without a ground-truth reference. 
However, these methods often rely on handcrafted features or are trained on limited image samples, which do not always align well with human perception.

To overcome this challenge, we leverage the advancements in Large Multimodal Models (LMMs) and introduce GScore, a new metric for assessing the quality of drag edited images. 
These large models, equipped with a vast number of parameters and trained on Internet-scale vision and language data, are capable of processing and analyzing a wide variety of images. 
We utilize LMMs as evaluators, providing them with the edited image and the original input image as a reference. 
We prompt these models to rate the images based on their perceptual quality on a scale from 0 to 10, with higher scores indicating better quality.

In our experiments, we explored the use of both GPT-4V~\citep{openai2023gpt4} and Gemini~\citep{team2023gemini} as evaluation agents. 
We find that the output from Gemini is more reliable and closely aligned with human visual judgment. 
Therefore, we select Gemini as the primary evaluation agent for assessing the quality of edited images in our work.

\section{Experiments}
\subsection{Implementation Details}
In our experiments, we use Stable Diffusion 1.5~\citep{rombach2022high} as the base model. 
For the optimization process, we employ the Adam optimizer~\citep{kingma2017adam} with a learning rate of 0.02. 
Before initiating the DDIM inversion, we finetune the diffusion model using LoRA with a rank of 16.
For the diffusion process, we set the number of denoising steps to $T_\text{max}=50$ and the inversion strength to 0.75, resulting in $T = 50 \times 0.75=38$. 
We do not utilize any text prompt for the diffusion model.
The features used in Eq.~\ref{eq:new_motion_loss} are extracted from the last layer of the U-Net.
In the AlDD framework, the radii for motion supervision (Eq.~\ref{eq:new_motion_loss}) and point tracking (Eq.~\ref{eqn:point_tracking}) are set to $r_1=4$ and $r_2=12$, respectively.
The drag size in Eq.~\ref{eq:new_motion_loss} is set to $\beta=4$, and the mask loss weight is set to $\lambda=0.2$.
The total number of drag operations is set to $K=70$, with $B=10$ drag operations per denoising step, resulting in $K/B=7$ denoising steps during the alternating phase.
For each drag operation, the number of motion supervision steps is $J=3$ in Eq.~\ref{eq:new_grad_descent}.
To enhance the editing performance, the Latent-MasaCtrl mechanism~\citep{cao2023masactrl} is incorporated starting from the 10th layer of the U-Net.

\begin{figure}[!ht]
\centering
\setlength{\tabcolsep}{1pt} 
\begin{tabular}{ccc}
\footnotesize User Edit & \footnotesize Ours & \footnotesize DragGAN  \\
\subfloat{\includegraphics[width=0.32\linewidth]{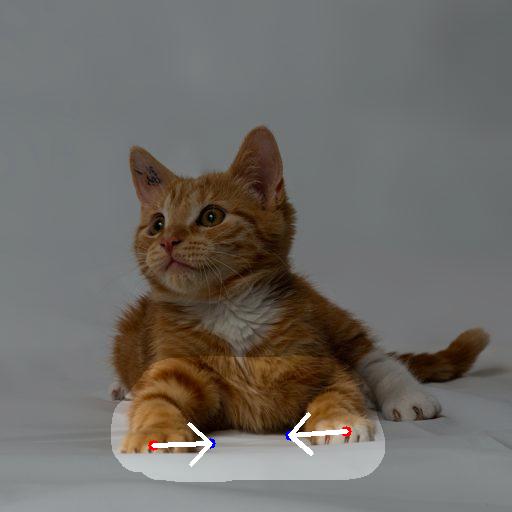}} &
\subfloat{\includegraphics[width=0.32\linewidth]{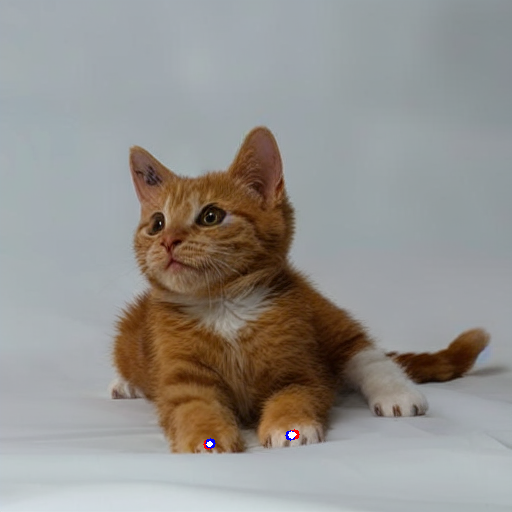}} &
\subfloat{\includegraphics[width=0.32\linewidth]{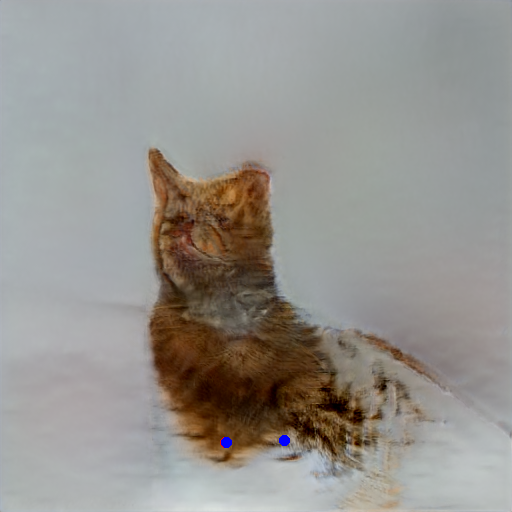}} 
 \\
\subfloat{\includegraphics[width=0.32\linewidth]{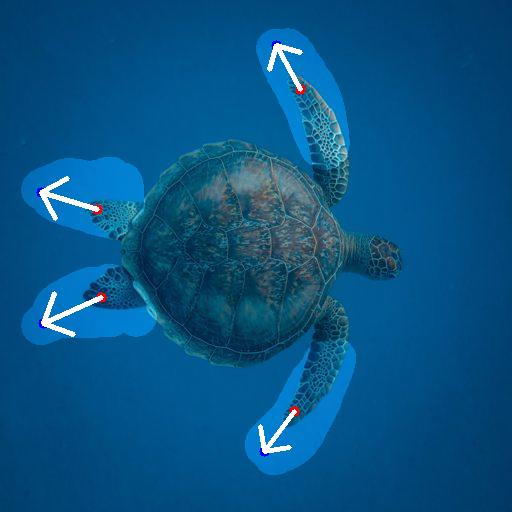}} &
\subfloat{\includegraphics[width=0.32\linewidth]{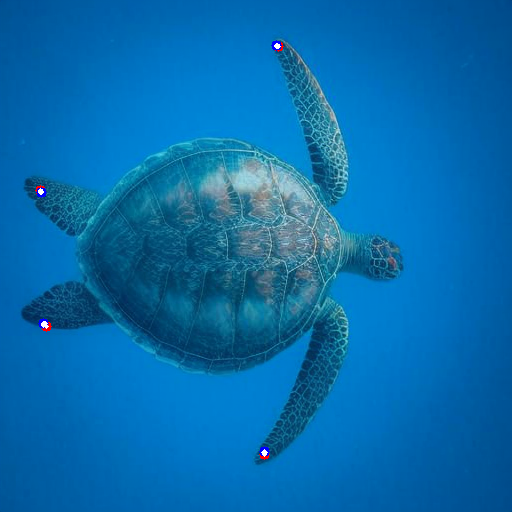}} &
\subfloat{\includegraphics[width=0.32\linewidth]{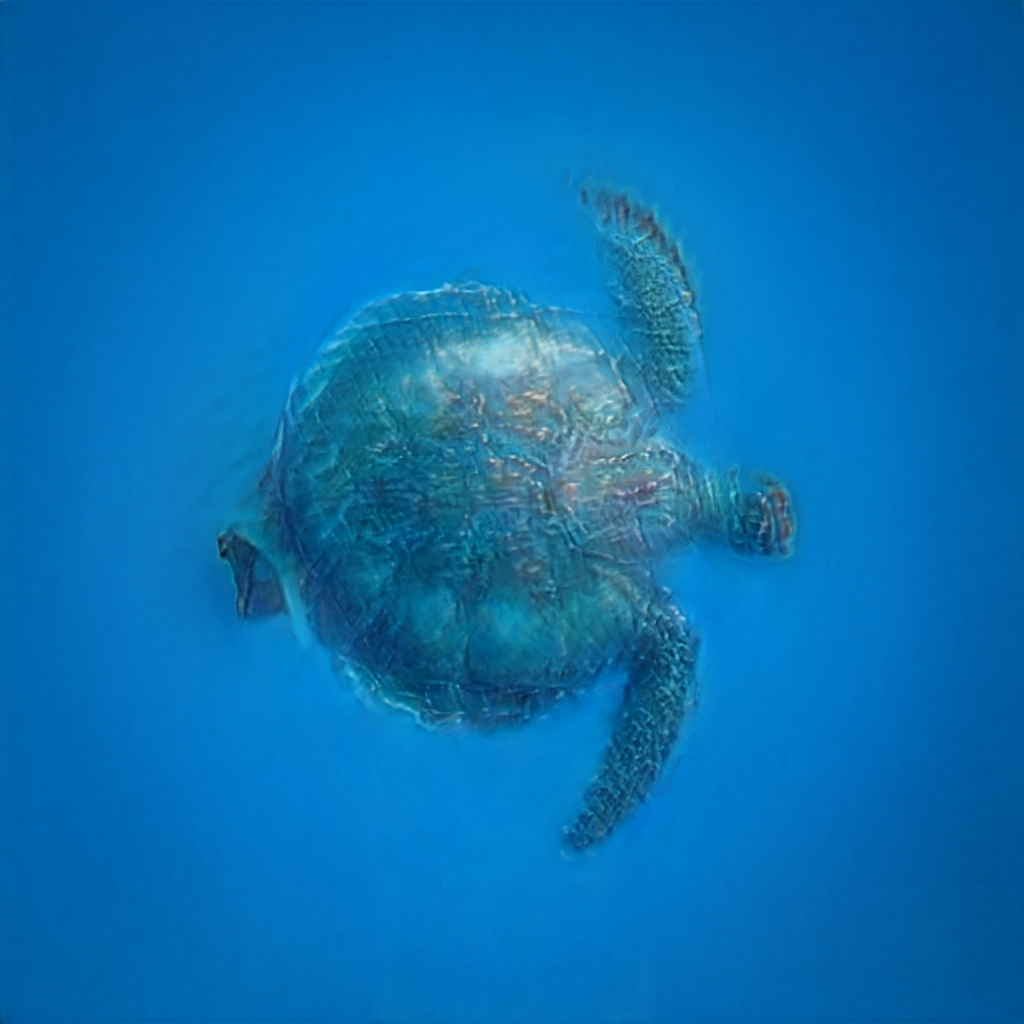}} 
\end{tabular}
\caption{Comparison with DragGAN~\cite{pan2023_DragGAN}. PTI~\cite{roich2022pivotal} is used in DragGAN for better GAN inversion. Our proposed method effectively edits the input images according to the specified control points, while DragGAN exhibits notable artifacts and low fidelity.}
\label{fig:draggan comparision}
\end{figure}

\begin{figure*}[t]
	\centering
	\setlength{\tabcolsep}{1pt}
	\begin{tabular}{ccccccccc}
		\footnotesize User Edit & \footnotesize Ours & \footnotesize DragDiffusion & \footnotesize SDE-Drag & ~ & \footnotesize User Edit & \footnotesize Ours & \footnotesize DragDiffusion & \footnotesize SDE-Drag \\
		\includegraphics[width=0.12\linewidth]{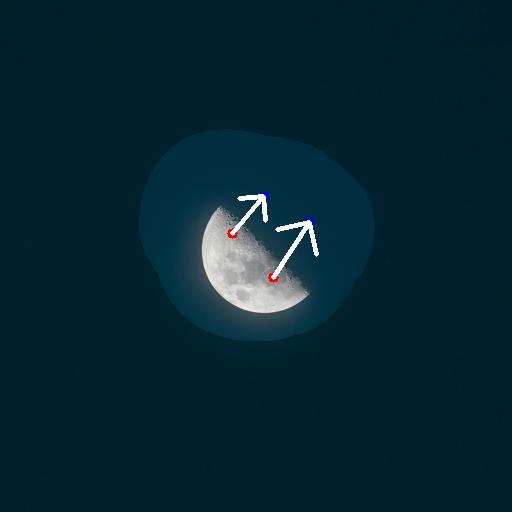} &
		\includegraphics[width=0.12\linewidth]{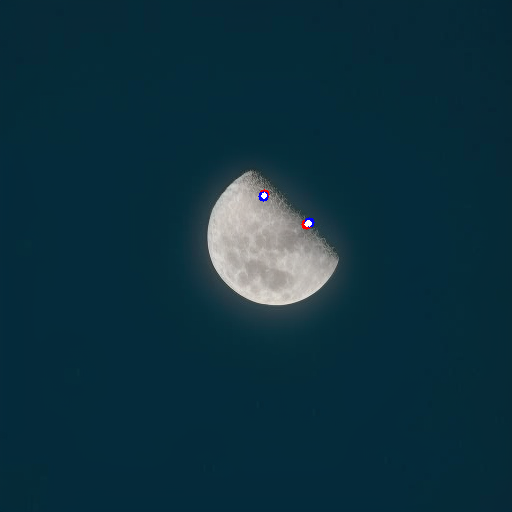} &
		\includegraphics[width=0.12\linewidth]{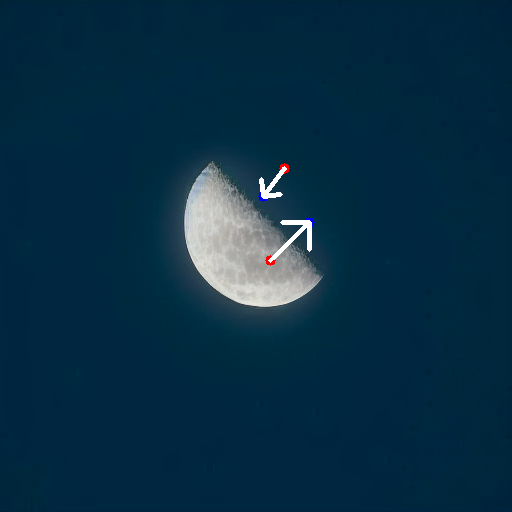} &
		\includegraphics[width=0.12\linewidth]{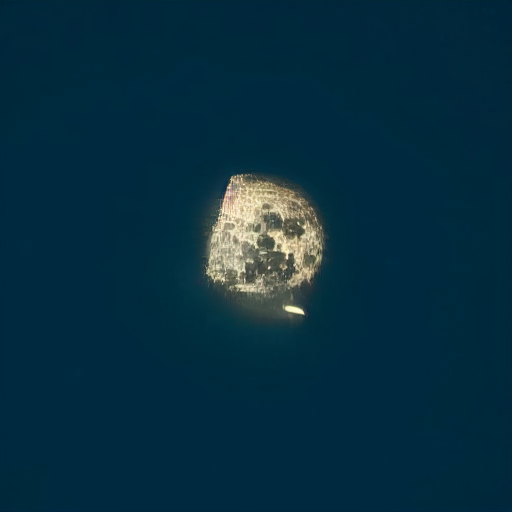} & &
		\includegraphics[width=0.12\linewidth]{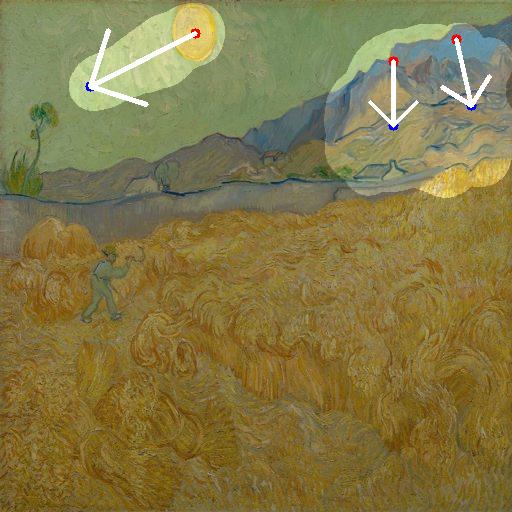} &
		\includegraphics[width=0.12\linewidth]{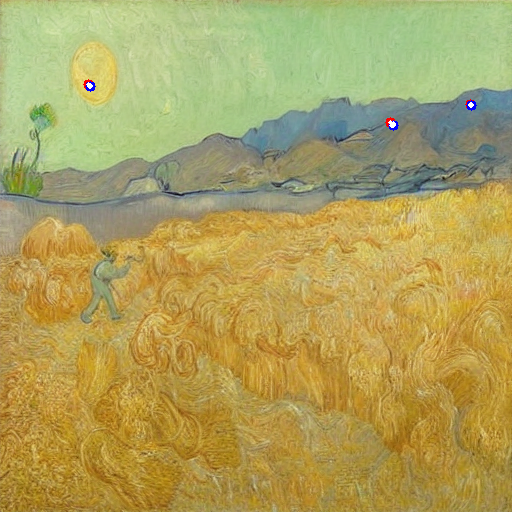} &
		\includegraphics[width=0.12\linewidth]{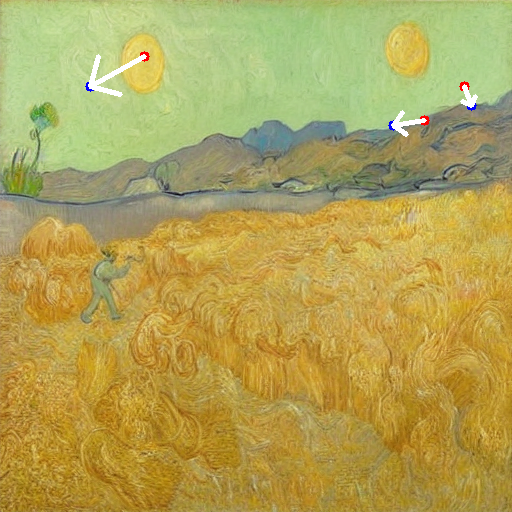} &
		\includegraphics[width=0.12\linewidth]{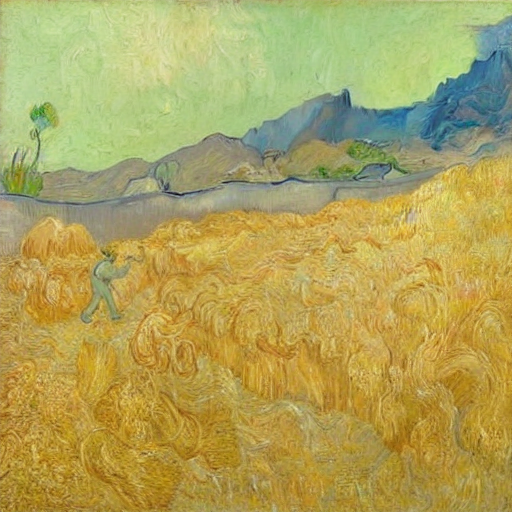} \\
		\includegraphics[width=0.12\linewidth]{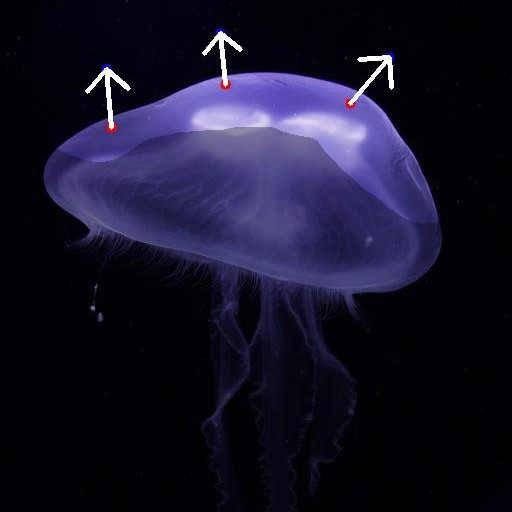} &
		\includegraphics[width=0.12\linewidth]{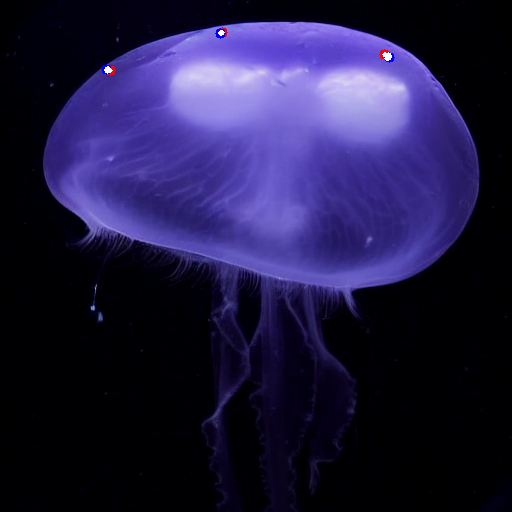} &
		\includegraphics[width=0.12\linewidth]{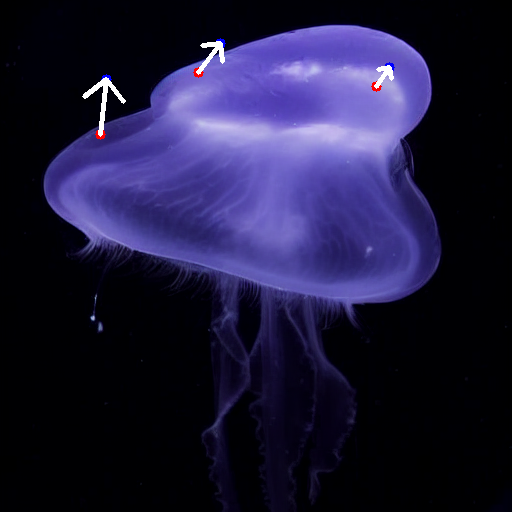} &
		\includegraphics[width=0.12\linewidth]{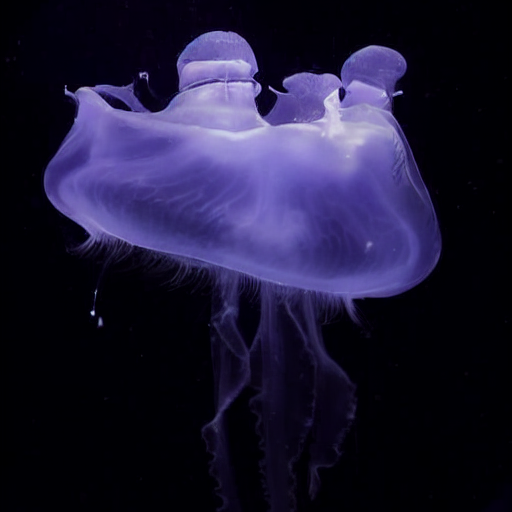} & &
		\includegraphics[width=0.12\linewidth]{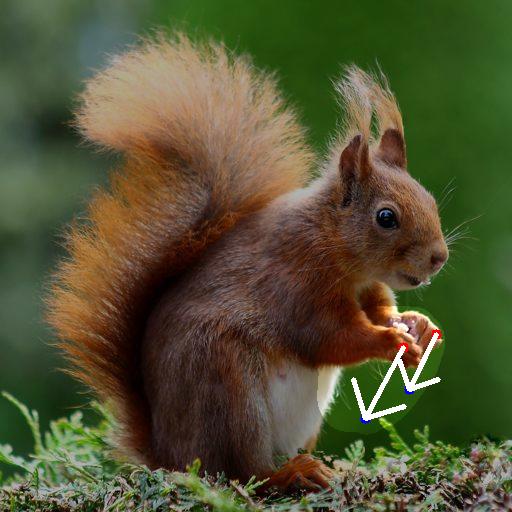} &
		\includegraphics[width=0.12\linewidth]{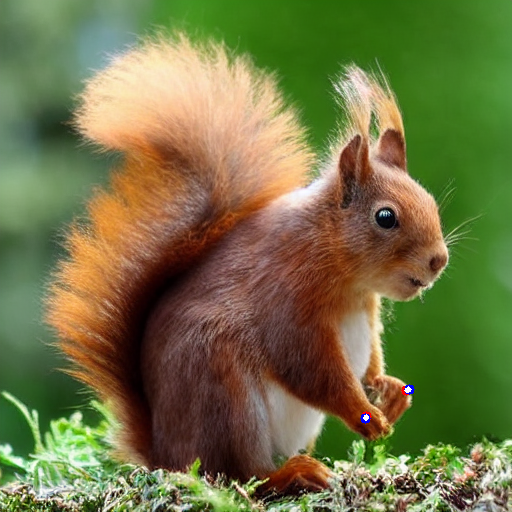} &
		\includegraphics[width=0.12\linewidth]{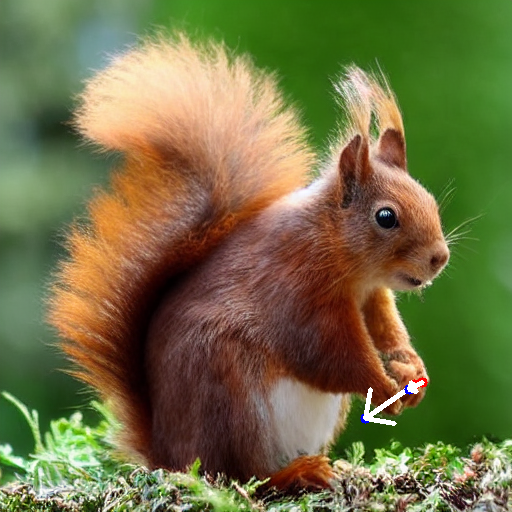} &
		\includegraphics[width=0.12\linewidth]{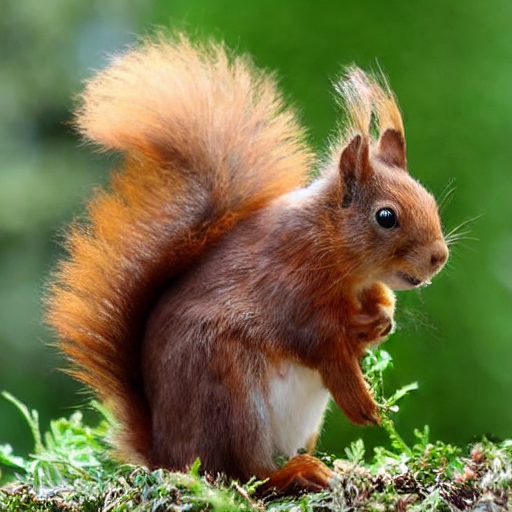} \\
		\includegraphics[width=0.12\linewidth]{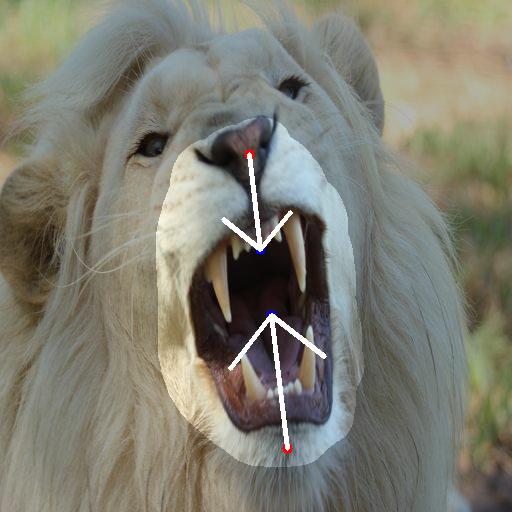} &
		\includegraphics[width=0.12\linewidth]{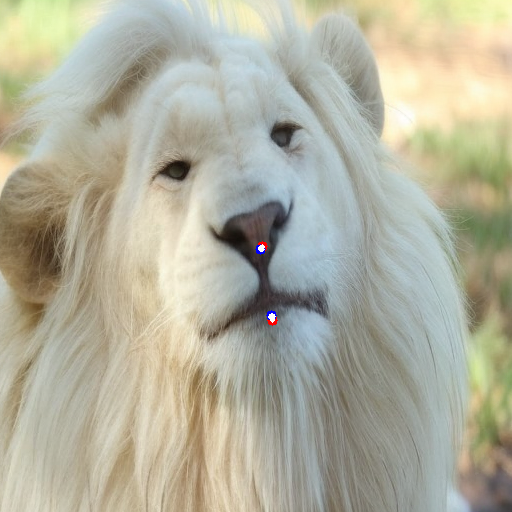} &
		\includegraphics[width=0.12\linewidth]{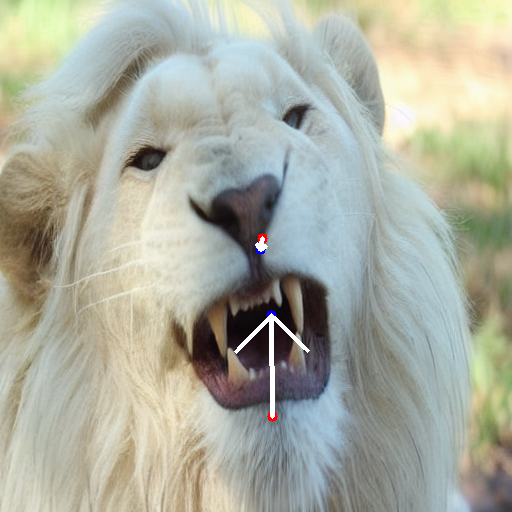} &
		\includegraphics[width=0.12\linewidth]{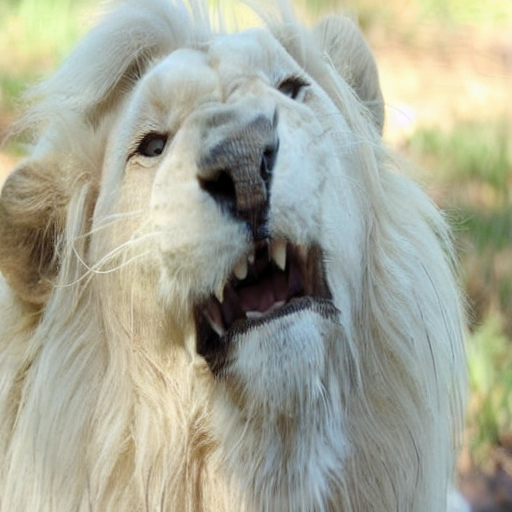} & &
		\includegraphics[width=0.12\linewidth]{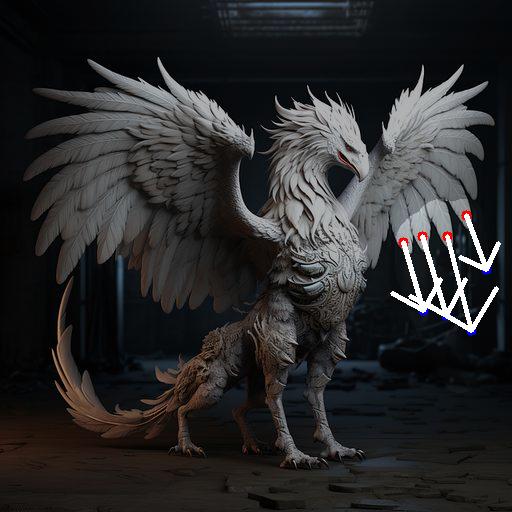} &
		\includegraphics[width=0.12\linewidth]{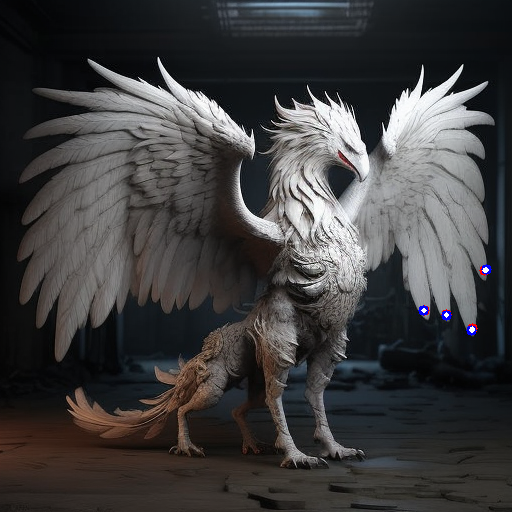} &
		\includegraphics[width=0.12\linewidth]{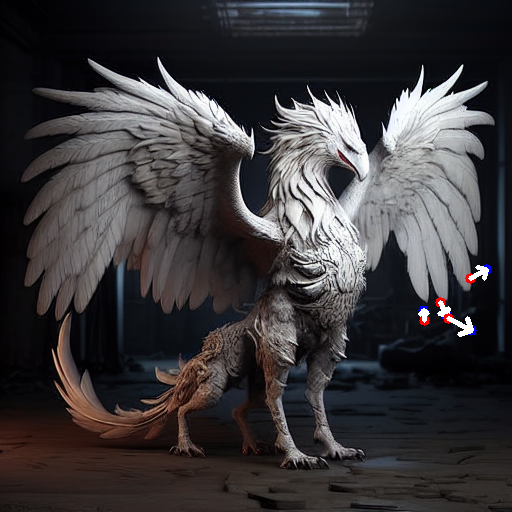} &
		\includegraphics[width=0.12\linewidth]{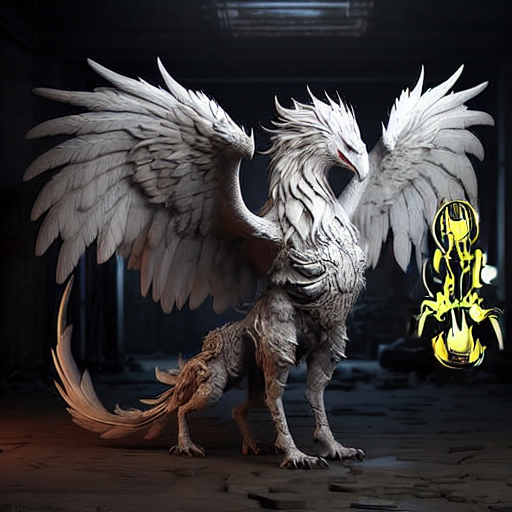} \\
		\includegraphics[width=0.12\linewidth]{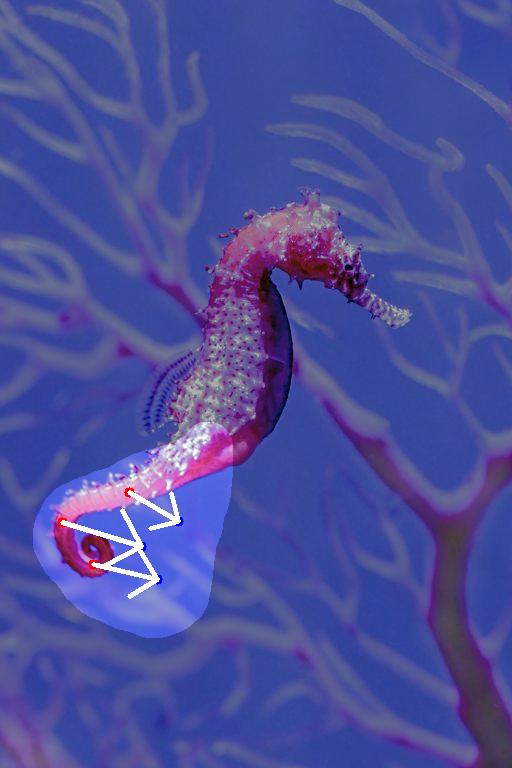} &
		\includegraphics[width=0.12\linewidth]{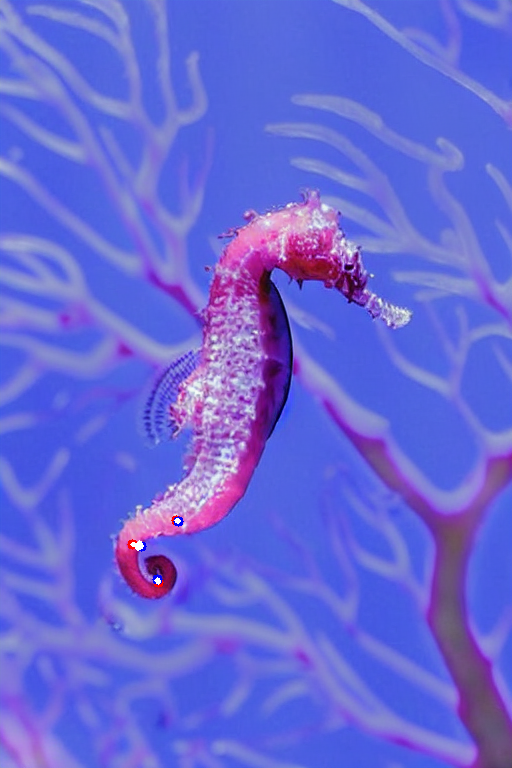} &
		\includegraphics[width=0.12\linewidth]{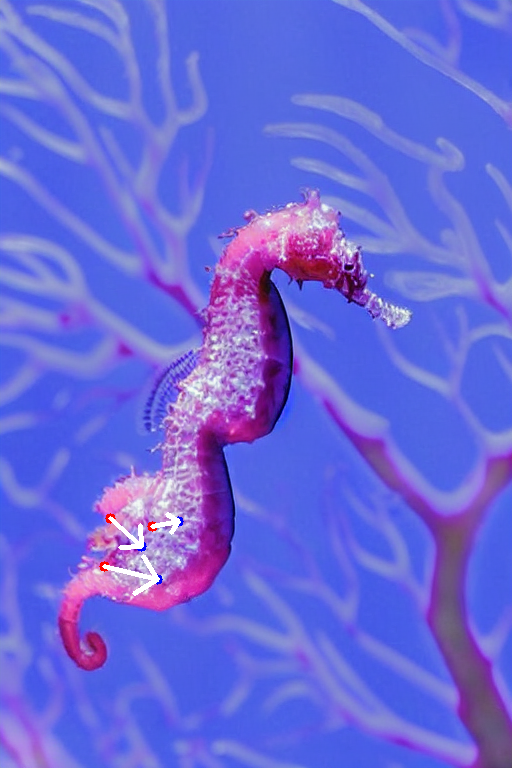} &
		\includegraphics[width=0.12\linewidth]{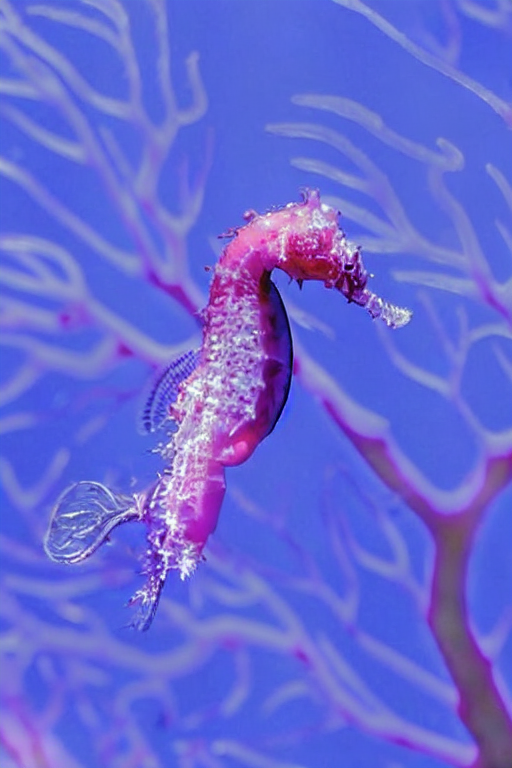} & &
		\includegraphics[width=0.12\linewidth]{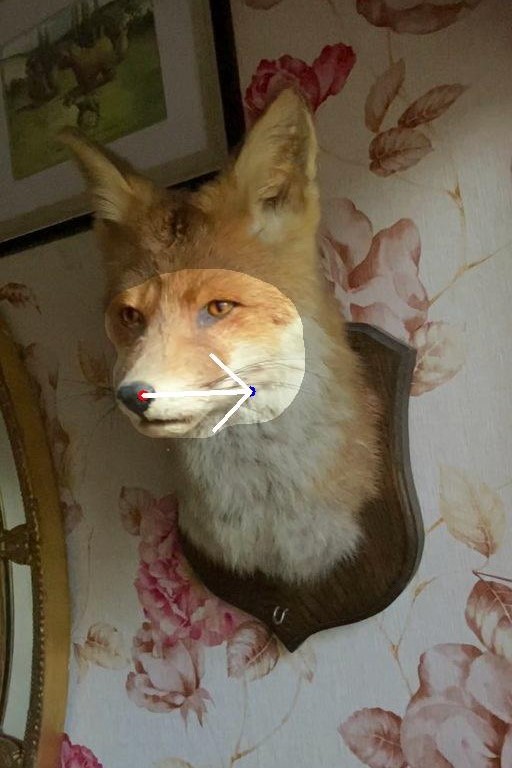} &
		\includegraphics[width=0.12\linewidth]{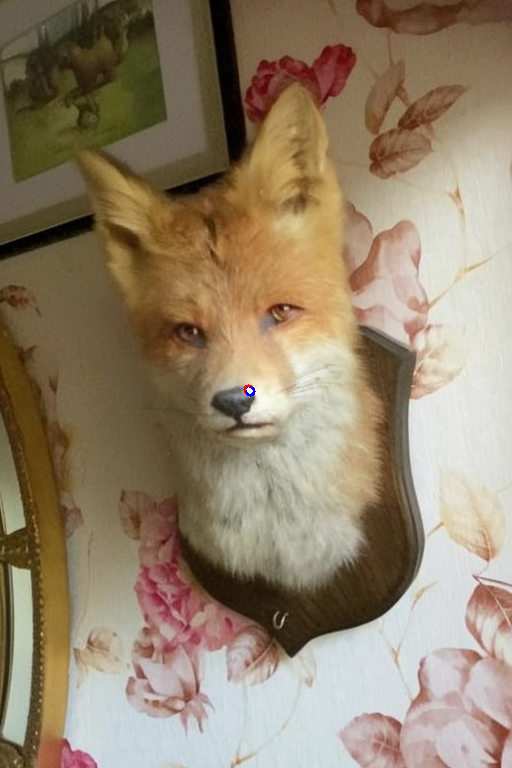} &
		\includegraphics[width=0.12\linewidth]{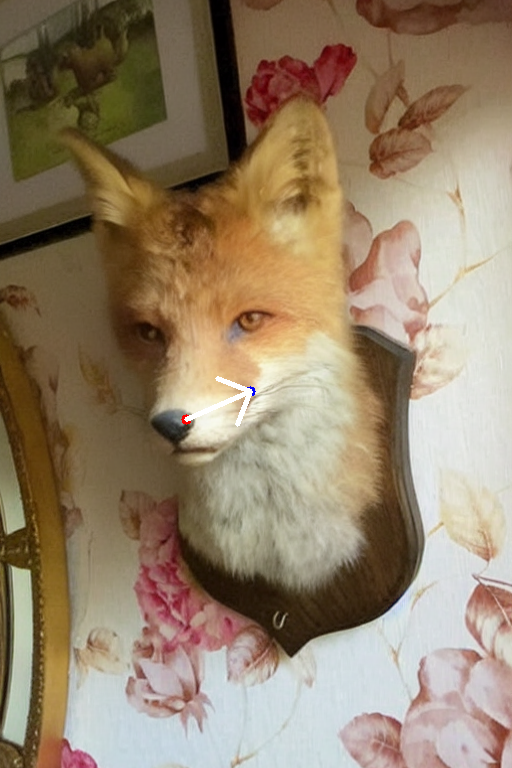} &
		\includegraphics[width=0.12\linewidth]{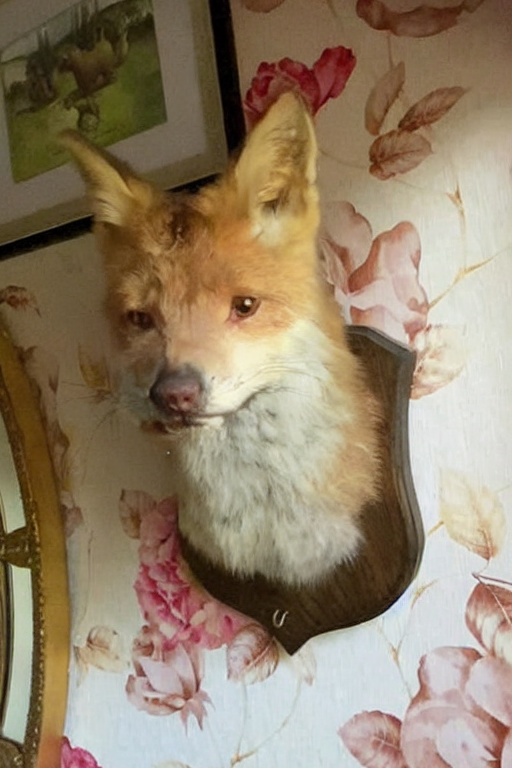}
	\end{tabular}
	\caption{Comparison with diffusion-based drag editing methods~\citep{shi2023dragdiffusion,nie2023blessing}. The proposed GoodDrag compares favorably against the baseline approaches in terms of both perceptual quality and accuracy of point movement.}
	\label{fig:dragdif_sde_comparision}
\end{figure*}

\begin{figure}[!ht]
\centering
\setlength{\tabcolsep}{1pt} 
\begin{tabular}{cccc}
\footnotesize User Edit & \footnotesize Ours & \footnotesize DragDiffusion & \footnotesize SDE-Drag \\
\subfloat{\includegraphics[width=0.24\linewidth]{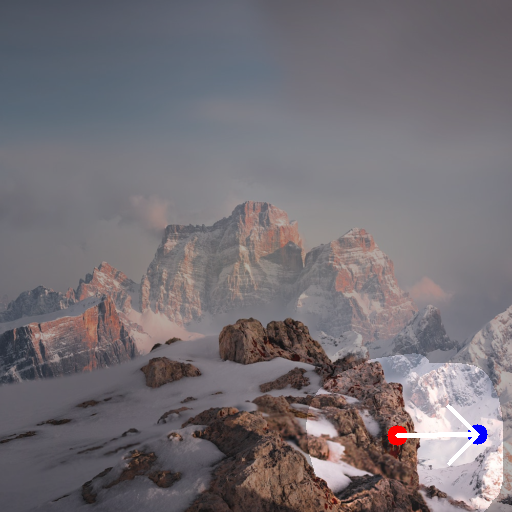}} &
\subfloat{\includegraphics[width=0.24\linewidth]{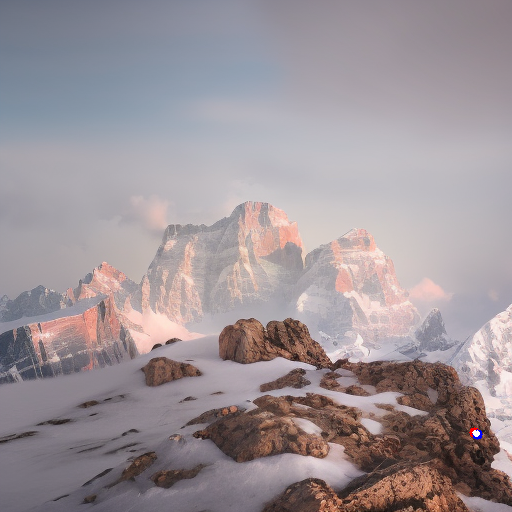}} &
\subfloat{\includegraphics[width=0.24\linewidth]{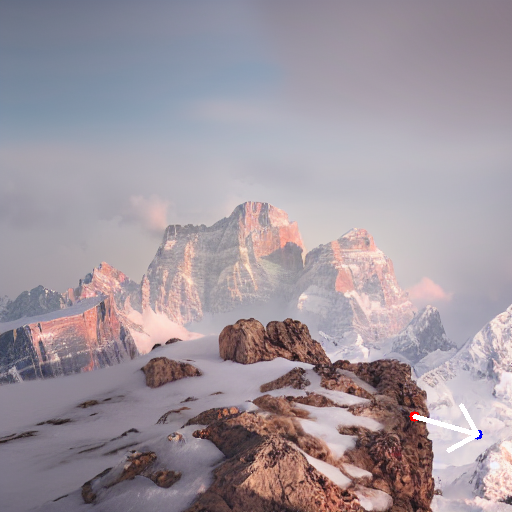}} &
\subfloat{\includegraphics[width=0.24\linewidth]{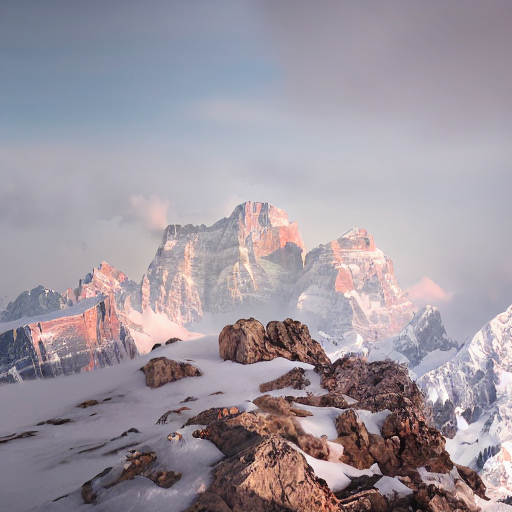}} \\
\subfloat{\includegraphics[width=0.24\linewidth]{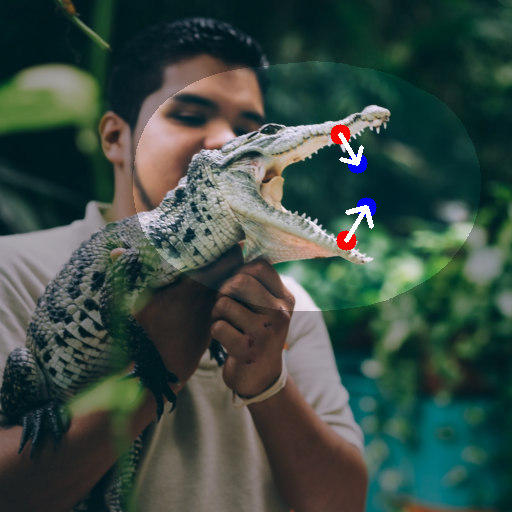}} &
\subfloat{\includegraphics[width=0.24\linewidth]{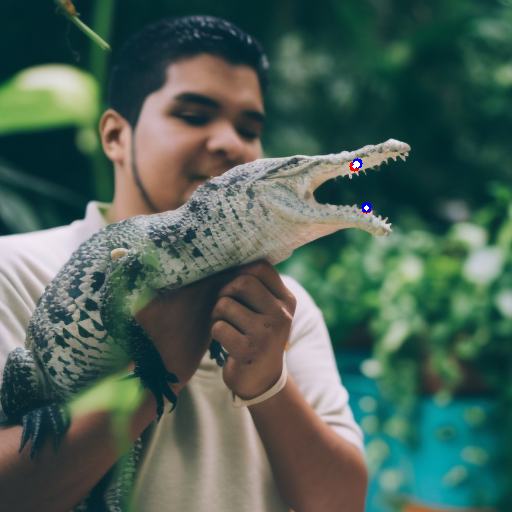}} &
\subfloat{\includegraphics[width=0.24\linewidth]{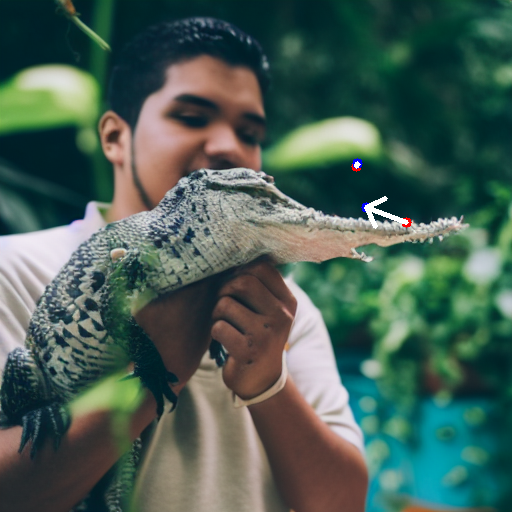}} &
\subfloat{\includegraphics[width=0.24\linewidth]{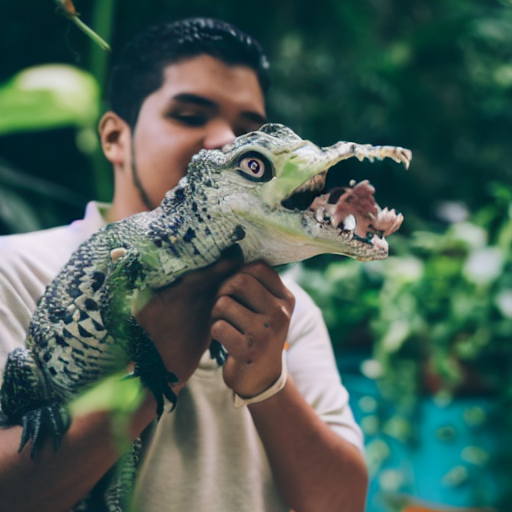}} \\
\subfloat{\includegraphics[width=0.24\linewidth]{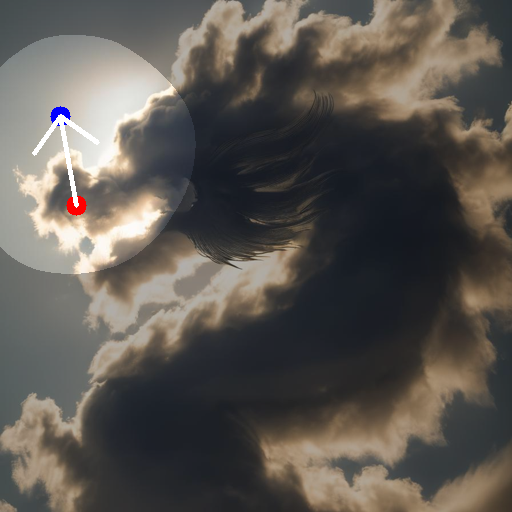}} &
\subfloat{\includegraphics[width=0.24\linewidth]{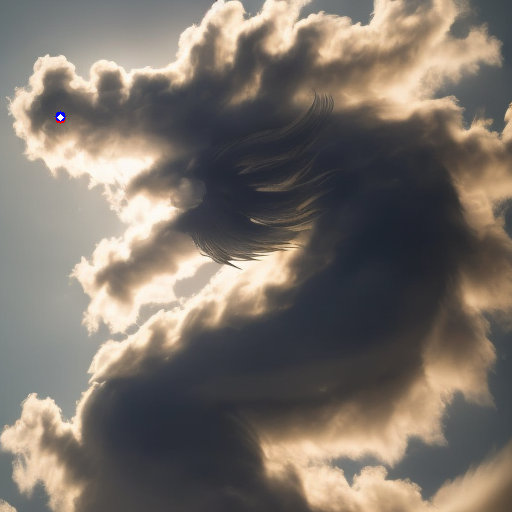}} &
\subfloat{\includegraphics[width=0.24\linewidth]{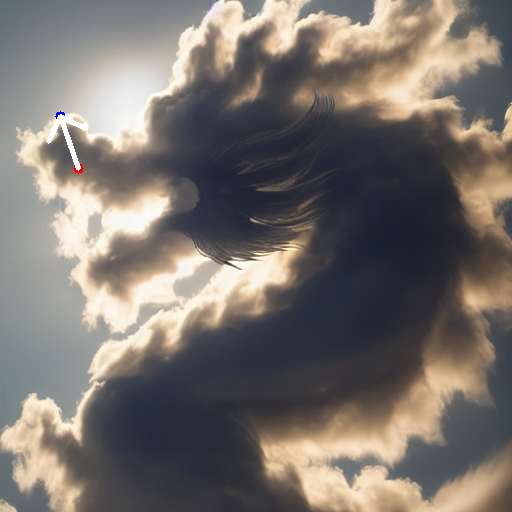}} &
\subfloat{\includegraphics[width=0.24\linewidth]{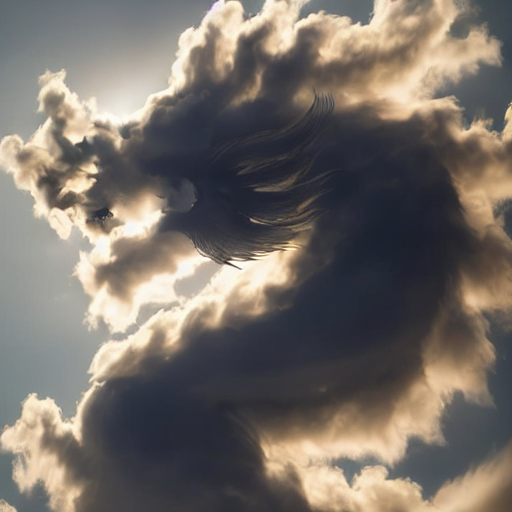}} \\
\subfloat{\includegraphics[width=0.24\linewidth]{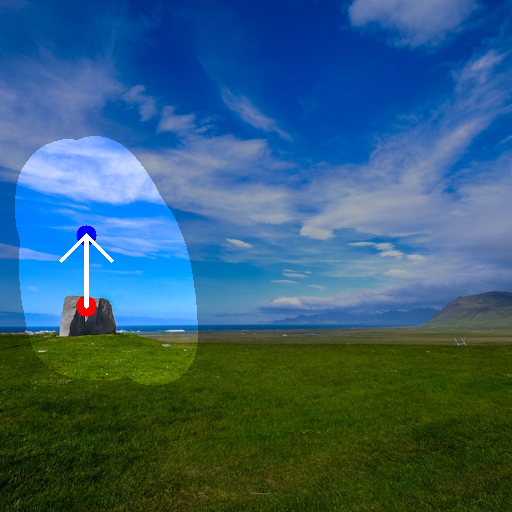}} &
\subfloat{\includegraphics[width=0.24\linewidth]{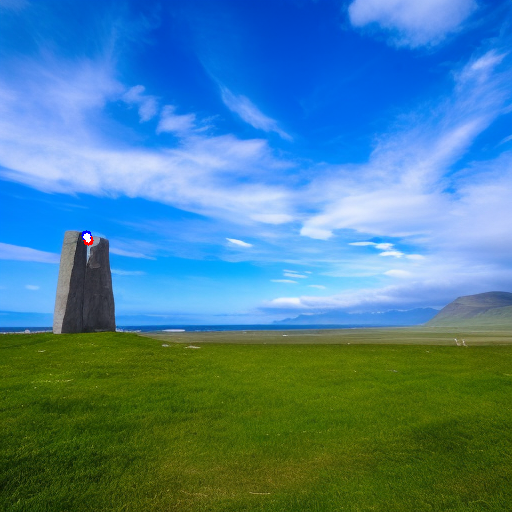}} &
\subfloat{\includegraphics[width=0.24\linewidth]{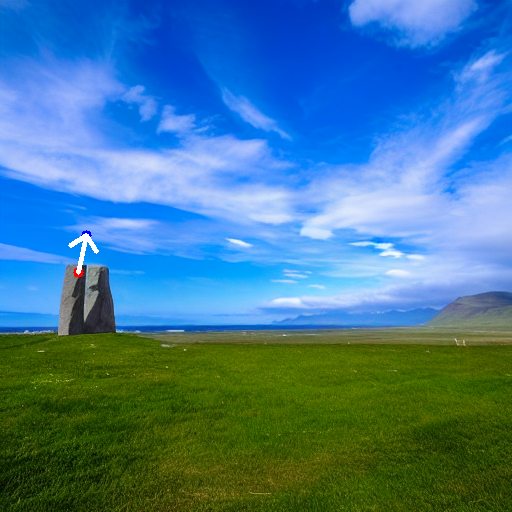}} &
\subfloat{\includegraphics[width=0.24\linewidth]{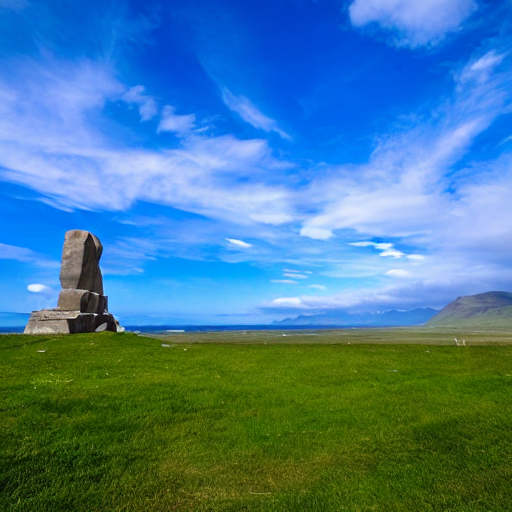}}
\end{tabular}
\caption{Comparison on images from \citep{shi2023dragdiffusion,nie2023blessing}. Note that these images do not have indication masks. For a fair comparison, we manually label masks for these images and apply the same masks across all methods.}
\label{fig:dataset comparision}
\end{figure}

\subsection{Comparison with SOTA}\label{sec:compare_sota}

\noindent\textbf{Qualitative evaluation.}
We first evaluate the proposed GoodDrag against DragGAN~\citep{pan2023_DragGAN} in Fig.~\ref{fig:draggan comparision}. 
The proposed method is able to effectively edit the input images according to the designated control points, whereas DragGAN suffers from notable artifacts and low fidelity. 
This superior performance is primarily due to the enhanced generative capabilities of diffusion models~\citep{dhariwal2021diffusion,rombach2022high} compared to GANs~\citep{karras2019style}, which enables GoodDrag to generalize well across various inputs.
Aside from the limited generative capability, DragGAN is also notably time-consuming. It requires finetuning a StyleGAN using PTI~\citep{roich2022pivotal} for better GAN inversion, which leads to significant computational overhead.

Next, we compare our method with diffusion-based approaches, including DragDiffusion~\citep{shi2023dragdiffusion} and SDE-Drag~\citep{nie2023blessing}. 
As shown in Fig.~\ref{fig:dragdif_sde_comparision} and \ref{fig:dataset comparision}, DragDiffusion has difficulty in accurately tracking the handling points and often fails to move semantic contents to the designated target locations.
On the other hand, while SDE-Drag achieves better point movement, it could introduce severe artifacts, resulting in low-fidelity images and unrealistic details. 
In contrast, GoodDrag demonstrates a stronger capability to precisely drag contents to the specified control points, producing much higher-quality results.
Note that the images in Fig.~\ref{fig:dataset comparision} are from the datasets of DragDiffusion and SDE-Drag, which do not provide indication masks.
For a fair comparison, we manually label masks for these images and apply the same masks across all methods.

\vspace{1mm}
\noindent\textbf{Quantitative evaluation.}
The evaluation in terms of DAI is presented in Table~\ref{tab:DAI_comparison}. 
We vary the patch radius $\gamma$ within the range of 1 to 20. 
When $\gamma$ is set to 1, the comparison focuses precisely on the feature of the control point. 
As the patch size increases, the DAI encompasses more contextual pixels, providing a broader perspective on drag accuracy.

As shown in Table~\ref{tab:DAI_comparison}, the proposed GoodDrag consistently outperforms the baseline methods across all values of $\gamma$, indicating higher accuracy in dragging semantic contents to the target points.
Notably, DragDiffusion employs 80 drag operations, whereas GoodDrag utilizes 70. 
However, with $J=3$ motion supervision steps in each drag operation (Eq.~\ref{eq:new_grad_descent}), GoodDrag effectively employs 210 motion supervision steps in its pipeline. 
In contrast, DragDiffusion requires only one motion supervision step per drag operation. 
To investigate whether the superior performance of GoodDrag is attributable to the increased number of supervision steps, we introduce a variant of DragDiffusion, termed DragDiffusion*, which uses 210 dragging operations, matching the number of motion supervision steps in our method.
While this adjustment slightly improves the results of DragDiffusion*, it still falls short of GoodDrag by a significant margin, highlighting the effectiveness of the proposed algorithm.

\setlength{\tabcolsep}{1.2mm}
{
\begin{table}[t]
\caption{Quantitative evaluation of drag accuracy in terms of DAI on Drag100. $\gamma$ corresponds to the patch radius in Eq.~\ref{eq:new_grad_descent}. Lower values indicate more accurate drag editing.}
\centering
\begin{tabular}{lccccccc}
\toprule
Method         & $\gamma=1$ & $\gamma=5$  & $\gamma=10$ & $\gamma=20$ \\ \midrule
DragDiffusion     &  0.1477     & 0.1439  & 0.1298 & 0.1146 \\ 
DragDiffusion* & 0.1189 & 0.1101  & 0.0979 & 0.0924 \\ 
SDE-Drag           & 0.1571 & 0.1437  &0.1291 & 0.1143 \\
GoodDrag              & \textbf{0.0696} & \textbf{0.0673}  & \textbf{0.0642} & \textbf{0.0623} \\ \bottomrule
\end{tabular}
\label{tab:DAI_comparison}
\end{table}
}

In addition, to evaluate the naturalness and fidelity of the edited images, we use the GScore proposed in Section~\ref{sec:metrics}.
As shown in Table~\ref{tab:Gemini score}, our method achieves an average GScore of 7.94 on the Drag100 dataset, outperforming DragDiffusion and SDE-Drag by a clear margin.

\setlength{\tabcolsep}{7mm}{
\begin{table}[t]
\caption{Quantitative evaluation of image quality in terms of GScore on Drag100. The GScore is on a scale from 0 to 10, with higher scores indicating better quality.}
\centering
\begin{tabular}{lc}
\toprule
Method                 & GScore $\uparrow$ \\ \midrule
DragDiffusion          & 6.87 \\ 
DragDiffusion* & 6.90 \\ 
SDEDrag           & 5.38 \\
Ours              & \textbf{7.94}  \\ \bottomrule
\end{tabular}
\label{tab:Gemini score}
\end{table}
}

\begin{figure}[!t]
	\footnotesize
	\begin{center}
		\begin{tabular}{cc}
			\hspace{-9mm}
			\includegraphics[width=0.49\linewidth]{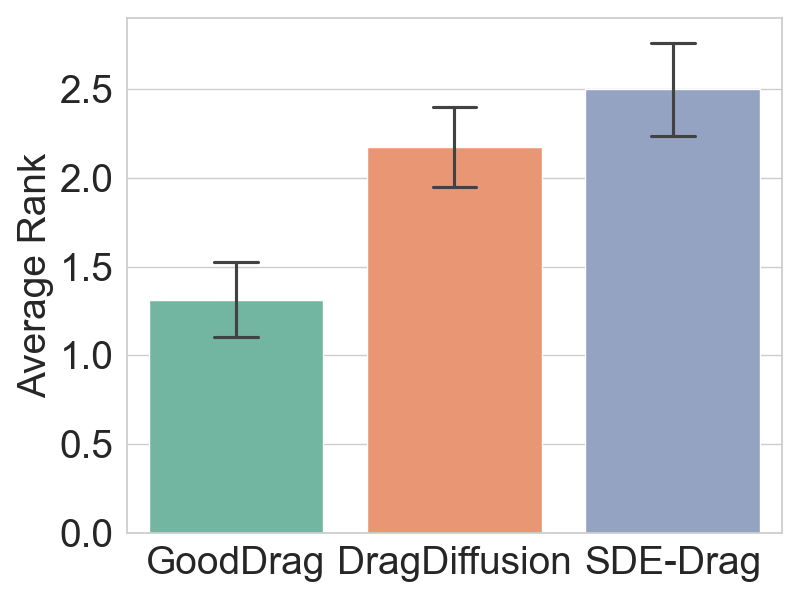} &
			\hspace{-10mm}
			\includegraphics[width=0.49\linewidth]{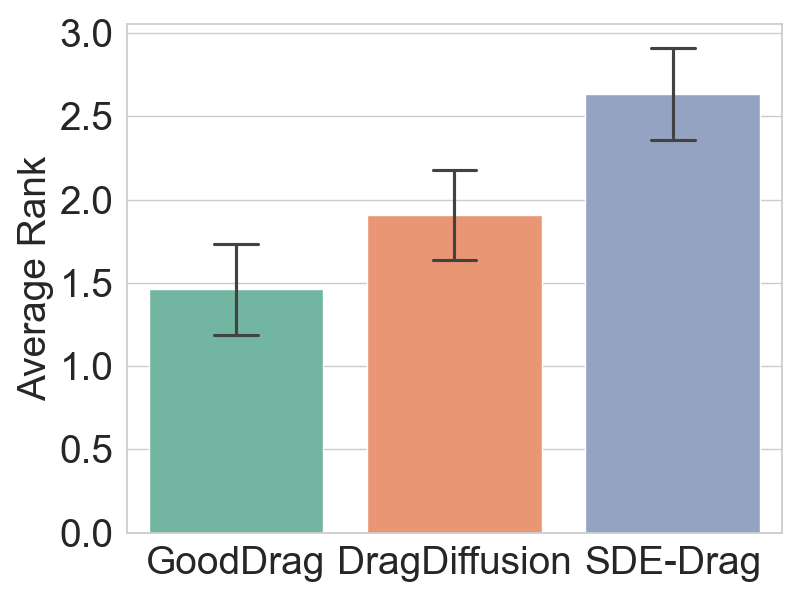}  \\
			\hspace{-4mm} \scriptsize (a) Drag accuracy &
			\hspace{-4mm} \scriptsize (b) Image quality \\
		\end{tabular}
	\end{center}
	\caption{User study on the drag accuracy (a) and perceptual quality (b) of the edited results. Lower ranks indicate better performance.} \label{fig:userstudy}
\end{figure}

\vspace{1mm}
\noindent\textbf{User study.}
For a more comprehensive evaluation of the drag editing algorithms, we conduct a user study with 12 images randomly selected from the Drag100 benchmark. 
Each image is processed by three different methods: DragDiffusion~\citep{shi2023dragdiffusion}, SDE-Drag~\citep{nie2023blessing}, and the proposed GoodDrag.
Subjects are asked to rank the edited results by each method with the input image as a reference (1 for the best and 3 for the worst).
The study is divided into two parts, with the ranking criteria being the accuracy of the drag editing and the perceptual quality of the results, respectively.
We receive responses from 27 participants, and the mean scores and standard deviations are presented in Fig.~\ref{fig:userstudy}. 
The proposed method is clearly preferred over other methods, suggesting its better capability in achieving precise drag editing (Fig.~\ref{fig:userstudy}(a)) while maintaining high perceptual quality (Fig.~\ref{fig:userstudy}(b)).

\subsection{Analysis and Discussion}\label{sec:analysis}

\begin{figure}[t]
	\footnotesize
	\begin{center}
		\begin{tabular}{ccc}
  \hspace{-7mm}  User Edit & \hspace{-14mm} w/o AlDD & \hspace{-14mm} w/ AlDD \\
			\hspace{-7mm}
			\includegraphics[width=0.32\linewidth]{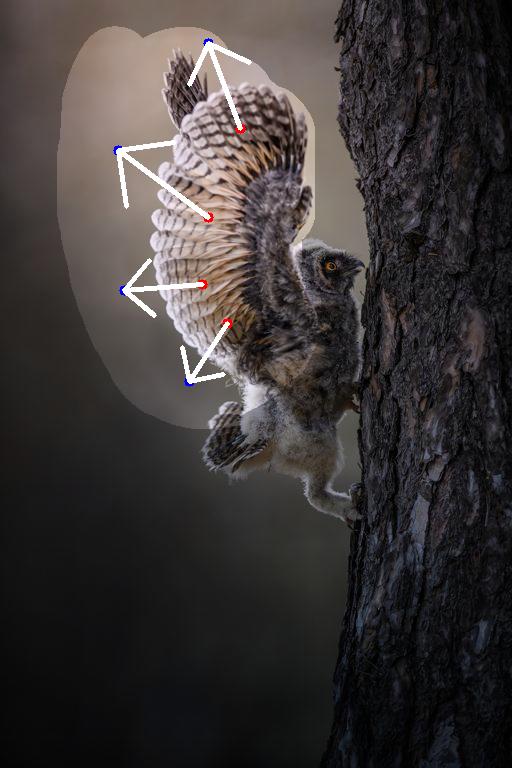} &
			\hspace{-14mm}
			\includegraphics[width=0.32\linewidth]{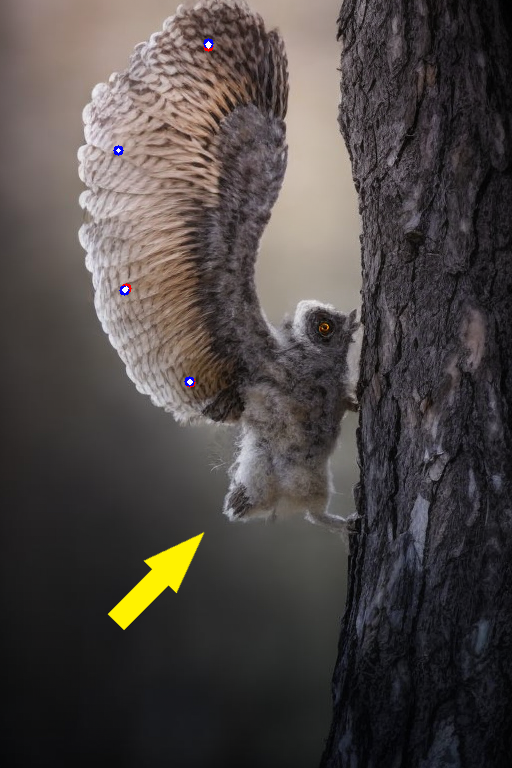} &
                \hspace{-14mm}
                \includegraphics[width=0.32\linewidth]{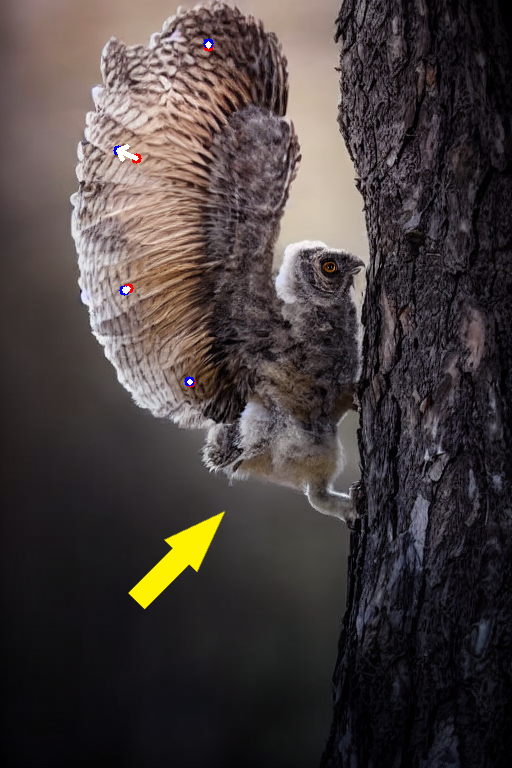} \\
 \hspace{-7mm}  10 Drags & \hspace{-14mm} 30 Drags & \hspace{-14mm} 50 Drags \\               
                \hspace{-7mm}
                \includegraphics[width=0.32\linewidth]{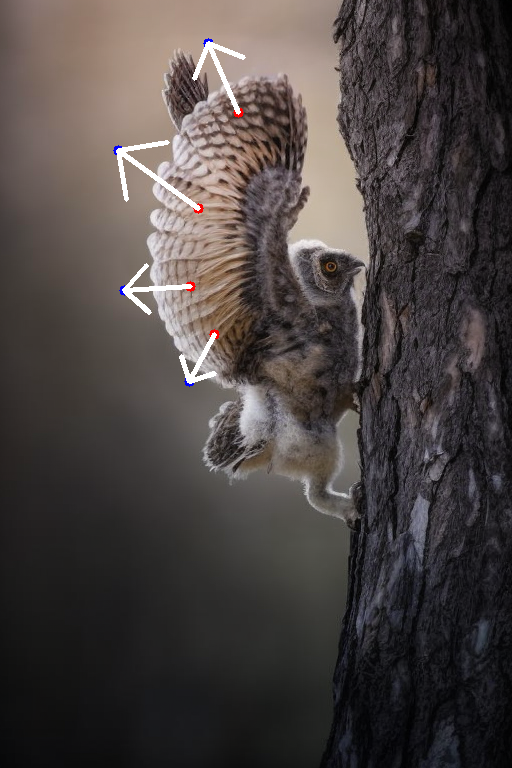} &
                \hspace{-14mm}
                \includegraphics[width=0.32\linewidth]{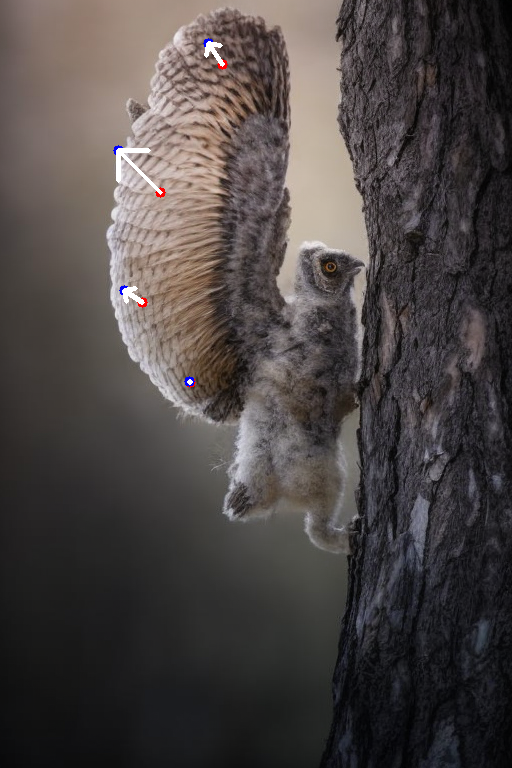} &
                \hspace{-14mm}
                \includegraphics[width=0.32\linewidth]{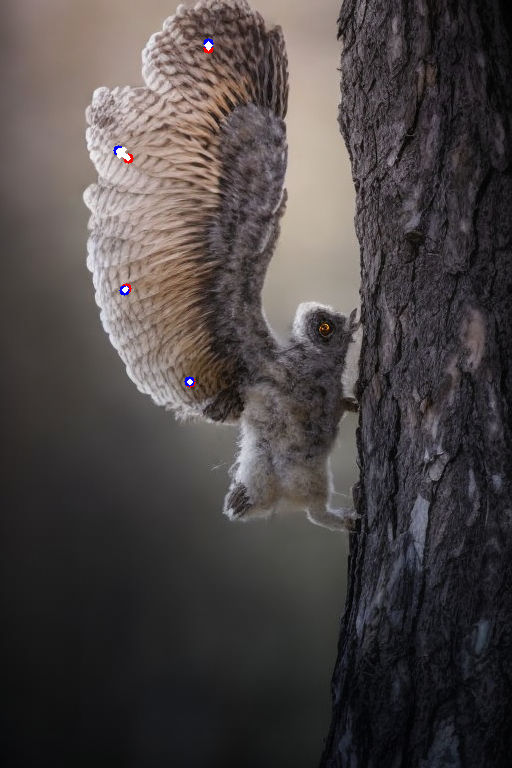} \\
		\end{tabular}
	\end{center}
	\caption{Effectiveness of AlDD.
In the first row, the result without AlDD shows noticeable inconsistencies in the owl's body compared to the input, while incorporating AlDD effectively addresses this issue. 
We use 70 drag operations by default. 
As shown in the second row, reducing the number of drag operations without AlDD improves fidelity but sacrifices the capability in relocating the semantic contents.
 } \label{fig:ablation_aldd}
\end{figure}

\begin{figure}
\footnotesize\hspace{-1em} (a) User Edit \hspace{2em} (b) w/o IP \hspace{2em} (c) w/ IP (Once) \hspace{2em} (d) w/ IP
\centering
\includegraphics[width=\linewidth]{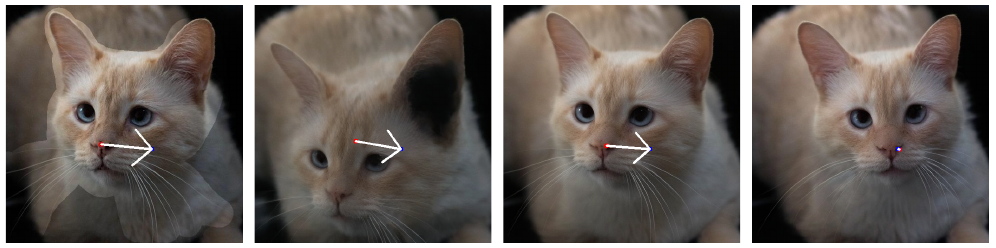}
\caption{The results without the proposed information-preserving motion supervision (IP) exhibit noticeable artifacts and dragging failures, as shown in (b), while incorporating IP effectively addresses this issue in (d). 
However, optimizing IP is inherently more challenging than the baseline approach, and directly using IP leads to inferior results in (c). To overcome this challenge, we propose employing multiple IP steps within a single drag operation, leading to the improved result in (d).}
\label{fig:drag multiple times}
\end{figure}

\noindent\textbf{Effectiveness of AlDD.}
As introduced in Section~\ref{sec:AlDD}, existing drag editing algorithms often suffer from low fidelity due to the accumulation of perturbations during the drag operations.
As shown in Fig.~\ref{fig:ablation_aldd}, the edited result without AlDD exhibits noticeable inconsistencies in the owl's body compared to the original image.
In contrast, incorporating AlDD significantly improves the fidelity of the edited result, ensuring that the owl's body remains faithful to the input image.

One might suggest that this fidelity issue could be mitigated by reducing the number of drag operations.
However, as illustrated in the second row of Fig.~\ref{fig:ablation_aldd}, while this approach does improve fidelity, it compromises the effectiveness of the drag editing, failing to relocate the content to the desired target locations. 
This underscores the importance of AlDD in achieving a better balance between fidelity and effective drag editing.

\begin{figure*}[t]
    \centering
    \includegraphics[width=0.95\linewidth]{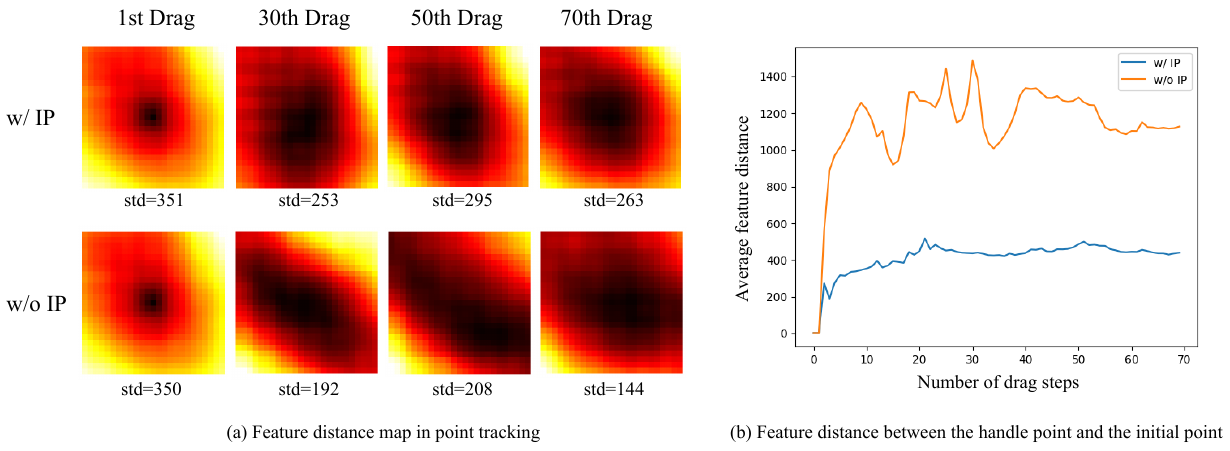}
    \caption{(a) shows the feature distance map from Eq.~\ref{eqn:point_tracking} at different drag steps. 
    More specifically, these heatmaps represent the feature distances between the original point $\boldsymbol{p}_i^0$ and the neighborhood of the current handle point ${\mathrm{\Omega}}(\boldsymbol{p}_i^k,r_2)$.
    The standard deviation (std) of the distances in each heatmap is provided below, where a small std indicates a diffused heatmap with indistinctive feature distances, and a large std indicates a more concentrated heatmap, resulting in generally more accurate localization of the smallest distance in Eq.~\ref{eqn:point_tracking}.
    (b) shows the feature distance between the handle point and the original point with the increase of drag steps. 
    The distance with the proposed information-preserving motion supervision (IP) is much smaller than that without IP, demonstrating its effectiveness in dealing with the feature drifting issue.
    }
    \label{fig:heatmap}
\end{figure*}

\vspace{1mm}
\noindent\textbf{Effectiveness of information-preserving motion supervision.}
As shown in Fig.~\ref{fig:drag multiple times}(b), the model without information-preserving motion supervision suffers from noticeable artifacts as well as dragging failures. 
In contrast, incorporating the information-preserving strategy effectively mitigates this issue, leading to improved results in Fig.~\ref{fig:drag multiple times}(d).

The feature distance between the handle point and the original point is shown in Fig.~\ref{fig:heatmap}(b),
where the proposed information-preserving motion supervision results in a substantially smaller feature distance (blue curve) compared to the model without this method (orange curve), underscoring its effectiveness in addressing feature drifting issues.

Furthermore, the information-preserving motion supervision also facilitates more accurate point tracking in Eq.~\ref{eqn:point_tracking}.
In Fig.~\ref{fig:heatmap}(a), we show the feature distance map between the original point $\boldsymbol{p}_i^0$ and the neighborhood of the current handle point ${\mathrm{\Omega}}(\boldsymbol{p}_i^k,r_2)$.
The heatmap with the information-preserving strategy is more concentrated with higher variance, thereby enabling more precise localization of the handle point. 
In contrast, the heatmap without this strategy is more diffused with lower variance.

Notably, adopting this information-preserving strategy presents challenges in the optimization of motion supervision due to the inherently larger feature distance in Eq.~\ref{eq:new_motion_loss} compared to Eq.~\ref{eq:base_motion_loss}.
This increased complexity can impede the movement of the handle point, as shown in Fig.~\ref{fig:drag multiple times}(c), where the cat's face remains stationary.
To overcome this issue, we employ multiple motion supervision steps within a single drag operation. 
As depicted in Fig.~\ref{fig:drag multiple times}(d), this approach effectively resolves the above issue, enabling the cat's face dragged to the desired orientation.

\setlength{\tabcolsep}{4.5mm}{
\begin{table}[t]
\caption{Correlations between various image quality assessment metrics and human visual perception.}
\centering
\footnotesize
\begin{tabular}{cccccc}
\toprule
      & TReS   & MUSIQ & TOPIQ & GScore\\ \midrule
$\rho \uparrow $      & 0.250   &  -0.125 & 0.083 & \textbf{0.708}  \\ 
 \bottomrule
\end{tabular}
\label{tab:spearman}
\end{table}
}

\vspace{1mm}
\noindent\textbf{Effectiveness of GScore.}
We compare various image quality assessment metrics, including TReS~\citep{golestaneh2022no}, MUSIQ~\citep{ke2021musiq}, TOPIQ~\citep{chen2023topiq}, and our proposed GScore, in terms of their alignment with human visual perception. 
We utilize the image quality rankings from the user study in Section~\ref{sec:compare_sota} and measure the correlation between these human rankings and the rankings produced by each metric.

Specifically, for the set of $N_s = 12$ images used in the user study, each image is processed by $N_m = 3$ different methods. For the $i$-th image, the human-assigned rankings for its $N_m$ results are denoted as $\{U_{ij}\}_{j=1}^{N_m}$, where $U_{ij}$ represents the rank assigned to the result of the $j$-th method. 
The rankings produced by an assessment metric for the same edited results are denoted as $\{R_{ij}\}_{j=1}^{N_m}$.
The correlation between a metric and the human judgment is defined as:
\begin{equation}
\rho = \frac{1}{N_s} \sum_{i=1}^{N_s} \rho_i,
\end{equation}
where $\rho_i$ is the Spearman's rank correlation coefficient~\citep{gauthier2001detecting} for the $i$-th image, calculated as:
\begin{equation}
\rho_i = 1 - \frac{6\sum_{j=1}^{N_m}(U_{ij}-R_{ij})^2}{N_m(N_m^2-1)}.
\end{equation}

The average correlations are presented in Table~\ref{tab:spearman}. While TReS, MUSIQ, and TOPIQ exhibit low (or even negative) correlations, GScore demonstrates a much higher correlation with the human visual system, indicating the effectiveness of GScore for assessing the perceptual quality of drag editing results.

\vspace{1mm}
\noindent\textbf{Runtime and GPU memory.}
We evaluate the runtime and GPU memory usage of GoodDrag with an A100 GPU. 
For an input image of size 512$\times$512, the LoRA phase takes approximately 17 seconds, while the remaining editing steps require about one minute. 
The total GPU memory consumption during this process is less than 13GB.

\section{Concluding Remarks}
In this work, we introduce GoodDrag, a method that enhances the stability and quality of drag editing. 
Leveraging our AlDD framework, we effectively mitigate distortions and enhance image fidelity by distributing drag operations across multiple diffusion denoising steps. 
In addition, we introduce information-preserving motion supervision to tackle the feature drifting issue, thereby reducing artifacts and enabling more precise control over handle points. 
Furthermore, we present the Drag100 dataset and two dedicated evaluation metrics, DAI and GScore, to facilitate a more comprehensive benchmarking of the progress in drag editing.
The simplicity and efficacy of GoodDrag establish a strong baseline for the development of more sophisticated drag editing algorithms. Future directions include exploring the integration of GoodDrag with other image editing tasks and extending its capabilities to video editing scenarios.

{
    \small
    \bibliographystyle{ieeenat_fullname}
    \bibliography{mybib}
}

\end{document}